\documentclass[10pt,journal,compsoc]{IEEEtran}
\ifCLASSOPTIONcompsoc
  \usepackage[nocompress]{cite}
\else
  \usepackage{cite}
\fi
\ifCLASSINFOpdf
\else
\fi
\usepackage{amsmath,amsfonts,amsthm}
\usepackage{algorithmic}
\usepackage{algorithm}
\usepackage{array}
\usepackage[]{subfig}
\usepackage{textcomp}
\usepackage{stfloats}
\usepackage{url}
\usepackage{verbatim}
\usepackage{graphicx}
\usepackage{cite}
\usepackage{booktabs}
\usepackage{multirow}
\usepackage{hyperref}
\usepackage[capitalize]{cleveref}
\crefname{section}{Sec.}{Secs.}
\Crefname{section}{Section}{Sections}
\Crefname{table}{Table}{Tables}
\crefname{table}{Tab.}{Tabs.}

\begin{document}
%
\title{Enabling Homogeneous GNNs to Handle Heterogeneous Graphs via Relation Embedding}

%
%
%
%

\author{
  Junfu~Wang,
  Yuanfang~Guo,~\IEEEmembership{Senior~Member,~IEEE,}\\
  Liang~Yang,~Yunhong~Wang,~\IEEEmembership{Fellow,~IEEE}

  \IEEEcompsocitemizethanks{
    \IEEEcompsocthanksitem J. Wang, Y. Guo, and Y. Wang are with the School of Computer Science and Engineering, Beihang University, Beijing 100191, China.\\
    E-mail: \{wangjunfu, andyguo, yhwang\}@buaa.edu.cn.
    \IEEEcompsocthanksitem L. Yang is with the School of Artificial Intelligence, Hebei University of Technology, Tianjin 300401, China. E-mail: yangliang@vip.qq.com. 
  }
}

\markboth{Journal of \LaTeX\ Class Files,~Vol.~14, No.~8, August~2015}%
{Shell \MakeLowercase{\textit{et al.}}: Bare Demo of IEEEtran.cls for Computer Society Journals}
%



\IEEEtitleabstractindextext{%
\begin{abstract}
  Graph Neural Networks (GNNs) have been generalized to process the heterogeneous graphs by various approaches.
  Unfortunately, these approaches usually model the heterogeneity via various complicated modules. 
  This paper aims to propose a simple yet effective framework to assign adequate ability to the homogeneous GNNs to handle the heterogeneous graphs. 
  Specifically, we propose Relation Embedding based Graph Neural Network (RE-GNN), which employs only one parameter per relation to embed the importance of distinct types of relations and node-type-specific self-loop connections. 
  To optimize these relation embeddings and the model parameters simultaneously, a gradient scaling factor is proposed to constrain the embeddings to converge to suitable values. 
  Besides, we interpret the proposed RE-GNN from two perspectives, and theoretically demonstrate that our RE-GCN possesses more expressive power than GTN (which is a typical heterogeneous GNN, and it can generate meta-paths adaptively).
  Extensive experiments demonstrate that our RE-GNN can effectively and efficiently handle the heterogeneous graphs and can be applied to various homogeneous GNNs.
\end{abstract}

}

\maketitle

\IEEEdisplaynontitleabstractindextext

%
\IEEEpeerreviewmaketitle

\IEEEraisesectionheading{\section{Introduction}\label{sec:intro}}

%
%
%
%


\IEEEPARstart{G}{raph} Neural Networks (GNNs) have shown great expressive power in graph representation learning \cite{gcn,gat,graphsage,bi-gcn,cvpr_gnn_1}.
They have been widely adopted in various downstream applications, 
such as social network analysis \cite{gnn_app_social1,gnn_app_social2}, protein prediction \cite{gnn_app_biology1,gnn_app_biology2}, 
traffic prediction \cite{gnn_app_traffic1,gnn_app_traffic2}, drug discovery \cite{gnn_app_drug}, etc.

Unfortunately, the majorities of traditional GNNs are designed for homogeneous graphs, which disregard the variations in both the node type and edge type.
The potential of GNNs in handling the data with complex relations has not been fully exploited.
Currently, heterogeneous graph, a.k.a, heterogeneous information network, which possesses more than one type of nodes or edges, has attracted more attentions from researchers, due to its ability to represent data with more complicated relations.

Recently, several literatures have generalized GNNs to handle heterogeneous graphs.
Existing heterogeneous GNNs can be classified into two categories, based on their mechanisms for handling the heterogeneity.

\begin{figure}[t]
  \centering
  \subfloat[Heterogeneous graph] {
    \label{fig1:a}
    \includegraphics[width=0.43\columnwidth]{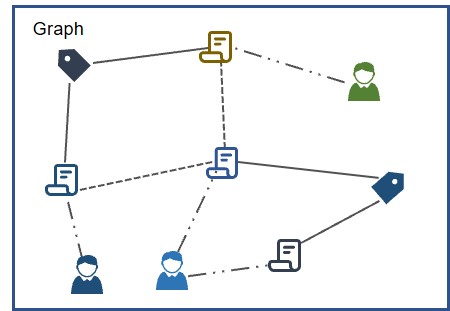}
  }
  \subfloat[Meta-path based homogeneous graphs] { 
    \label{fig1:b}
    \includegraphics[width=0.43\columnwidth]{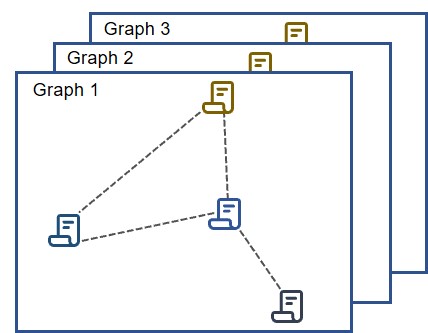}
  }

  \subfloat[Relation based subgraphs] { 
    \label{fig1:c}
    \includegraphics[width=0.43\columnwidth]{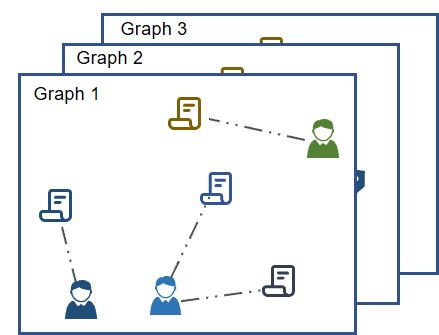}
  }
  \subfloat[Homogeneous graph with relation-specific weights] { 
    \label{fig1:d}
    \includegraphics[width=0.43\columnwidth]{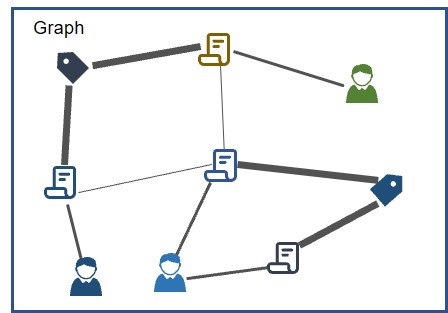}
  }
  \label{fig1}
  \caption{Different mechanisms to handle the heterogeneous graphs. A heterogeneous academic graph is utilized as an example.}
\end{figure}

The first type of approaches utilizes the composite relations, i.e., meta-paths, to convert the heterogeneous graph to several meta-path based homogeneous graphs, as shown in \cref{fig1:b}.
For example, in the ACM dataset, the papers, which are not directly connected in the original heterogeneous graph yet connected with the same authors, are connected via a composite relation (\textit{paper-author-paper}).
These meta-path based heterogeneous GNNs \cite{han,magnn,gtn,diffmg} utilize the handcrafted or learned meta-paths as shortcuts to boost the efficiency of the message passing among the target types of nodes.
However, they require certain prior knowledge to either design the meta-paths \cite{han,magnn} or generate the meta-paths with excessive computations \cite{gtn,diffmg}.
Besides, the performances of these methods is highly correlated to the quality of the constructed meta-paths.

The other kind of approaches directly models the heterogeneity, especially the heterogeneous relations.
They usually handle the subgraph of each type of relations (as shown in \cref{fig1:c}) separately, via constructing different modules \cite{rgcn,rgsn,hetgnn,hgt,r-hgnn}.
Since their complexities are correlated to the number of relation types, they are less efficient when there exist many types of relations.
Besides, \cite{rshn} generates the relation representations and utilizes them as a part of the neighbor messages, i.e., concatenate them with the node representations.
Then, the node attributes and relation representations are aggregated in a homogeneous manner.

Recently, \cite{hgb} indicates that the homogeneous GNNs possess more potential to handle the heterogeneous graphs than previously reported \cite{han,gtn,magnn}.
It firstly reduces the complexity of the heterogeneous graph by converting its nodes and edges to a single type.
Then, a specific homogeneous GNN, such as Graph Attention Network (GAT) \cite{gat}, can achieve a comparable performance with the well-designed feature initialization strategies on certain datasets.
Although \cite{hgb} then proposes an improvement to make GAT more suitable for processing heterogeneous graphs, it has not provided a general mechanism for handling heterogeneity with various homogeneous GNNs.
However, different GNNs \cite{gcn,gat,gin,graphsage,bi-gcn} usually possess unique advantages in handling homogeneous graphs with different properties, which may be suitable for different applications.
Thus, it is necessary to enable various homogeneous GNNs to process different heterogeneous graphs while preserving their advantages.

This paper aims to assign adequate abilities to the homogeneous GNNs for handling the heterogeneous graphs, by proposing a simple yet effective framework.
The key of our work is to effectively convert the entire to-be-processed heterogeneous graph to one homogeneous graph, i.e., to homogenize the various types of nodes and relations.
Generally, the attributes in different types of nodes are extracted from different perspectives.
To homogenize the graph, the features in different types of nodes are firstly projected into the same feature space, by assuming that different types of nodes are correlated implicitly.
For example, in a typical heterogeneous academic network, all types of nodes, e.g., papers, authors, conferences and subjects, are actually correlated to a certain extent.
Then, each type of nodes can pass messages according to various relations.
%
Considering the heterogeneity of relations, in the neighbourhood aggregation step, the messages from the nodes with different relations should not possess the same importance.
Therefore, we must determine the appropriate importance for each relation in the neighbourhood aggregation step.

Specifically, we propose Relation Embedding based Graph Neural Network (RE-GNN), where only one embedding parameter is employed for each relation to model the aggregation importance.
Then, the heterogeneous graph is adaptively converted to a weighted homogeneous graph by the proposed relation embeddings, as shown in Fig. \ref{fig1:d}.
Similar to the proposed relation embeddings, node-type-specific self-loop embeddings are exploited to add the self-loop connections.
Then, the weighted graph with self-loops can be directly processed by the traditional homogeneous GNNs, such as \cite{gcn,gat,gin}.
Since the heterogeneity is embedded into the weighted homogeneous graph, the importance of a neighbor in the aggregation step is highly correlated with the type of its relation to the current node. 
With this simple  yet effective framework, homogeneous GNNs can possess adequate abilities to handle the heterogeneous graphs.

To effectively model the heterogeneity, the learned weights of distinct relations are expected to be highly distinguishable.
However, since the numerical values of relation embeddings are much larger than the other parameters, a straightforward simultaneous optimization of the relation embeddings and the other parameters cannot fully exploit the potential of relation embeddings.
To tackle this \textit{numerical inconsistency}, a gradient scaling factor is proposed to enlarge/suppress the updating modifications to the embeddings in each iteration, which enables the relation embeddings to gradually converge to the optimal values, to generate effective weights for the relations.

The proposed RE-GNN can be interpreted from two perspectives.
Although RE-GNN is proposed from the perspective of spatial aggregations, it also possesses a meta-path based explanation.
Specifically, when utilizing GCN \cite{gcn} as backbone, we reveal that our RE-GCN can degenerate to the Graph Transformer Network (GTN) \cite{gtn}, a typical meta-path based heterogeneous GNN, which automatically generates composite relations.
Besides, we theoretically demonstrate that our RE-GCN possesses more expressive power than GTN, without explicitly generating the meta-paths.

Extensive experiments demonstrate that our RE-GNN can effectively and efficiently handle the heterogeneous graphs and can be applied to various homogeneous GNNs.

Our contributions are summarized as follows:
\begin{itemize}
  \item We propose a simple yet effective framework, named RE-GNN, to assign adequate ability to the homogeneous GNNs for handling heterogeneous graphs.
  \item To tackle the \textit{numerical inconsistency}, we propose a gradient scaling factor to effectively optimize the relation embeddings and other model parameters simultaneously.
  \item We interpret the proposed RE-GNN from two perspectives, and theoretically demonstrate that our RE-GCN possesses more expressive power than GTN, a typical heterogeneous GNN.
  \item Extensive experiments demonstrate that our RE-GNN can effectively and efficiently handle the heterogeneous graphs and can be easily applied to various homogeneous GNNs.
\end{itemize}

\section{Related Work}
\label{sec:relat}
\subsection{Graph Neural Networks}
Inspired by the traditional deep neural networks, Graph Neural Networks \cite{gnn} are designed to handle the irregular graph data.
Typical GNNs are usually designed from either the spectral or spatial perspectives.
Spectral methods employ the graph spectral theory \cite{gft} to define the graph convolution operation \cite{gcn1,gcn2,gcn,sgc}.
Spatial methods design the graph convolution operation directly on the graph, and aggregate the message from the spatial neighbors \cite{message-passing,graphsage,gat,fastgcn}.
Unfortunately, most of them are only designed to handle the homogeneous graphs.

\subsection{Heterogeneous GNNs}

To process the heterogeneous graph, researchers generalize traditional GNNs to heterogeneous graphs.
Currently, there exist two types of approaches to model the heterogeneity.

The first type of approaches is developed based on meta-path constructions, which is firstly introduced by HAN \cite{han}.
It converts a heterogeneous graph to multiple homogeneous graphs by various manually-designed meta-paths.
For each type of meta-paths, the meta-path based neighbors are connected in the corresponding homogeneous graph. 
Then, the results of each meta-path based homogeneous graph are fused by an attention scheme.
Based on the above mechanism, MAGNN \cite{magnn} utilizes the intermediate nodes along the meta-path, instead of only considering the meta-path based neighbors.
These two methods require certain prior domain knowledge to design the meta-paths for each heterogeneous graph.
GTN \cite{gtn} firstly generates meta-paths via a soft selection of edge types.
Then, an ensemble of GCNs is utilized to process the learned composite relations.
Subsequently, the neural architecture search (NAS) technique \cite{nas-survey} is utilized in DiffMG \cite{diffmg} to seek for a suitable meta-graph to model the more complicated composite relations.

The other type of approaches intends to model the heterogeneity directly, via different kinds of nodes and relations.
R-GCN \cite{rgcn} models the relations in knowledge graphs by employing specialized parameter matrices, which separately constructs GNNs on each relation graph.
R-GSN \cite{rgsn} further improves it by leveraging attention-based intra- and inter-relation aggregations.
HetGNN \cite{hetgnn} encodes node heterogeneous content and aggregates the neighbors from the same node type by using Bi-LSTMs, and combines the node embeddings from different node types with an attention scheme.
HGT \cite{hgt} utilizes a transformer network for each relation to model the importance of each relation.
Besides, RSHN \cite{rshn} constructs an edge-centric coarsened line graph to generate the relation representations, and then transfers the content of the node and relation representations to the target nodes, in the message passing process.
R-HGNN \cite{r-hgnn} learns the relation representations and fuse the relation-aware representations of the target nodes semantically.
All of these methods directly model the heterogeneity via complicated mechanism and possess excessive parameters.
HGB \cite{hgb} generalizes GAT \cite{gat} by adding the edge type attention to the original pair-wise self-attention mechanism.
However, it does not provide a general mechanism to enable the homogeneous GNNs to handle the heterogeneity.

\section{Preliminaries}
\label{sec:Prel}

\subsection{Heterogeneous Graph}

A graph is defined as $\mathcal{G} = (\mathcal{V}, \mathcal{E}, \mathcal{F}, \mathcal{R}, \phi, \varphi)$, where $\mathcal{V}$ and $\mathcal{E}$ stand for the collections of nodes and edges, respectively.
$\phi: \mathcal{V} \rightarrow \mathcal{F}$ and $\varphi: \mathcal{E} \rightarrow \mathcal {R}$ are the node type and edge type mapping functions, respectively.
$\mathcal{F} = \{\phi(v): \forall v \in \mathcal{V} \}$ represents the dynamic range of the node type projections and $\mathcal{R}=\{\varphi(e) : \forall e \in \mathcal{E}\}$ is the dynamic range of the edge type projections.
A typical heterogeneous graph \cite{heterogeneous-graph} contains more than one type of nodes or edges, i.e., $|\mathcal{F}| + |\mathcal{R}| > 2$.
$A_i$ denotes the corresponding adjacency matrix of the edge type $i\in\mathcal{R}$.
On the contrary, in homogeneous graphs, both the node and edge types only possess one valid value, i.e., $|\mathcal{F}| = |\mathcal{R}| = 1$.

\subsection{Meta-path}

Meta-path is widely adopted in the learning of heterogeneous graphs.
A meta-path models a path passing through multiple relations (which may belong to different types of relations), e.g., 
$v_1 \stackrel{r_1}{\longrightarrow} v_2 \stackrel{r_2}{\longrightarrow} ... \stackrel{r_l}{\longrightarrow} v_{l+1}$, where $v_i \in \mathcal{V}$ and $r_j \in \mathcal{R}$.
It describes a pair of nodes $v_1$ and $v_{l+1}$, which is connected by a composite relation.
Note that $R=r_1\odot r_2 \odot ... \odot r_l$, where $\odot$ denotes the composite operator on relations.
$v_1$ and $v_{l+1}$ are the meta-path based neighbors, and the corresponding composite adjacency matrix is $A_P=A_{r_1}A_{r_2}\dots A_{r_l}$.

\subsection{Graph Convolutional Network}

Graph Convolutional Network (GCN) \cite{gcn} has become the most popular GNN in the past few years.
Given a homogeneous graph $\mathcal{G}$, its graph convolution operation can be described as
\begin{equation}
H^{(l+1)} = \sigma(\tilde{A}H^{(l)}W^{(l)}),
\label{gcn-propagation}
\end{equation}
where $\tilde{A}=\hat{D}^{-\frac{1}{2}}\hat{A}\hat{D}^{-\frac{1}{2}}$ is the normalized adjacency matrix, $\hat{A}=A+I$ denotes the adjacency matrix with self-loops, and $\hat{D}$ represents the corresponding degree matrix.
$W^{(l)}\in\mathbb{R}^{d_{in}^{(l)}\times d_{out}^{(l)}}$ contains the learnable parameters.
$H^{(l+1)}$ is the output of the $l$-th layer and the input of the $(l+1)$-th layer, and $H^{(0)} = X$.
$\sigma$ stands for the non-linear activation function, e.g., ReLU.
For a directed graph (i.e., asymmetric adjacency matrix), $\hat{A}$ can be normalized by the inverse of the degree matrix, i.e., $\hat{D}^{-1}$, as $\tilde{A}=\hat{D}^{-1}\hat{A}$.

\subsection{Graph Transformer Network}
\label{sec-gtn}

Graph Transformer Network (GTN) \cite{gtn} generates meta-paths in an end-to-end manner, which possesses the ability to search for the task-specific meta-paths from all the possible meta-paths.
Since GTN is a typical meta-path based heterogeneous GNN and it will be compared to our approach in latter sections, we give a brief review here.

Firstly, GTN learns an adjacency matrix of $l$-length meta-paths via a GT layer
\begin{equation}
    A_P = \left( \sum_{r_1\in \mathcal{R}} \alpha_{r_1}^{(1)}A_{r_1} \right) \cdots \left( \sum_{r_l\in \mathcal{R}} \alpha_{r_l}^{(l)}A_{r_l} \right),
    \label{eq-gtn-1}
\end{equation}
where $\alpha_{r_i}^{(j)}={\rm softmax}_{j}\left(w_{jr_i}\right)$ is the soft weight for the edge type $r_i\in{\mathcal R}$ and $w$ is the learnable parameters.

Then, the learned adjacency matrix is fed into the standard GCN as
\begin{equation}
    Z^{(l)} = \sigma(\tilde{D}_P^{-1}\tilde{A}_P^{(l)}H^{(l)}W^{(l)}),
    \label{eq-gtn-2}
\end{equation}
where $\tilde{A}_P^{(l)}=A_P^{(l)} + I$ and $\tilde{D}_P$ denotes the corresponding degree matrix.

At last, the representations learned by multiple generated meta-paths are concatenated as
\begin{equation}
    H^{(l+1)} = \Arrowvert_{i=1}^C Z^{(l)}_i,
    \label{eq-gtn-3}
\end{equation} 
where $\Arrowvert$ denotes the concatenation operator and $C$ represents the number of generated meta-paths.
This architecture can be viewed as an ensemble of GCNs on multiple generated meta-path relations.

\begin{figure*}[t]
    \centering
    \includegraphics[width=1.8\columnwidth]{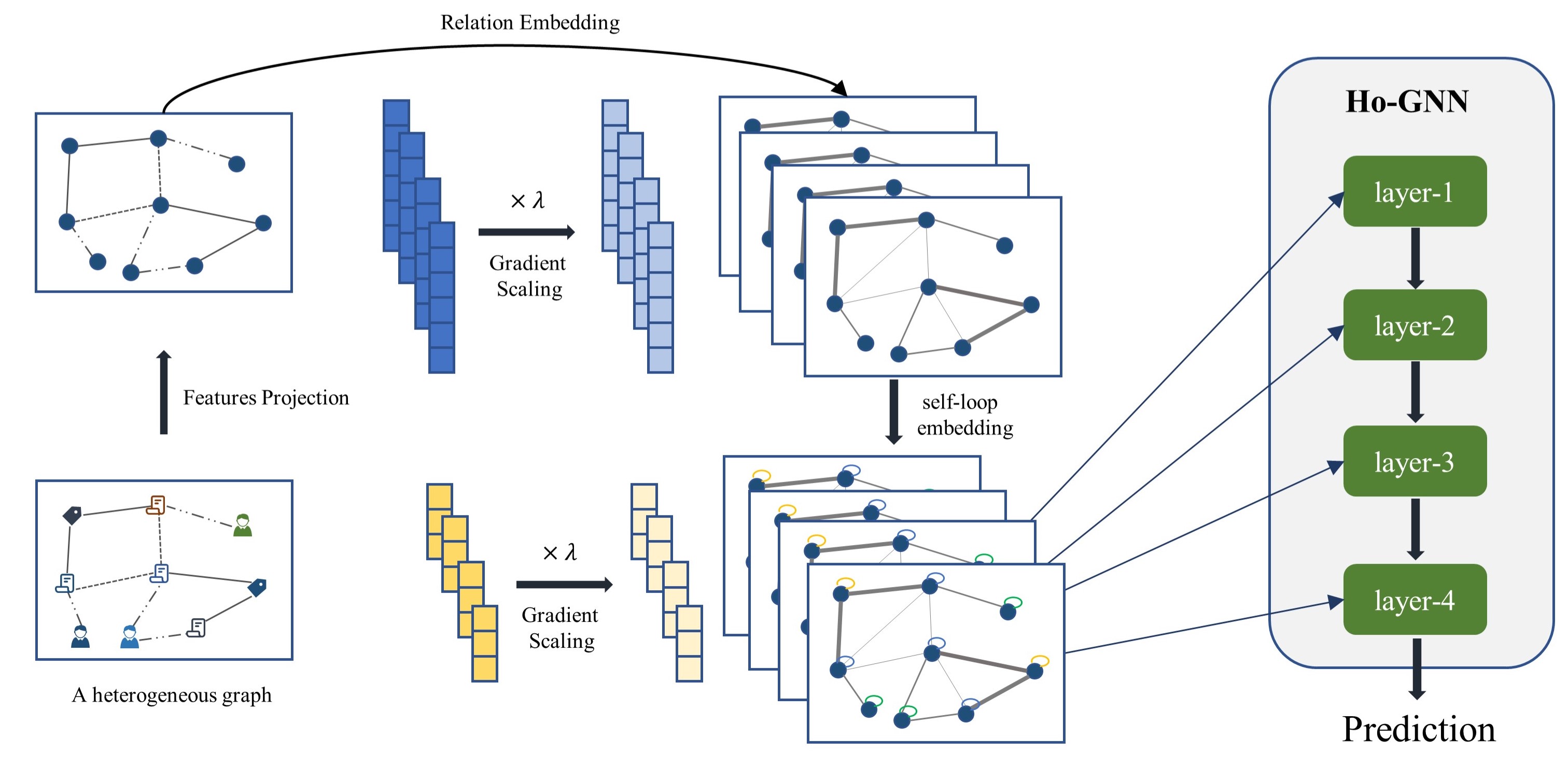}
    \caption{The illustration of the Relation Embedding based Graph Neural Network (RE-GNN).}
    \label{fig2-framework}
    \vspace{-0.4cm}
\end{figure*}

\section{Methodology}
\label{sec:method}


To assign adequate abilities to the homogeneous GNNs for handling the heterogeneous graphs, in this section, we propose a simple yet effective framework, named Relation Embedding based Graph Neural Network (RE-GNN).
Specifically, by assuming that all types of nodes are associated in particular relations, the node features are projected via a node-type-specific transformation matrix.
Then, for the heterogeneous topology, one embedding parameter per relation is exploited to learn the importance of different types of relations and self-loop connections.
Then, the heterogeneous graph can be converted to a weighted homogeneous graph, which can be handled by typical homogeneous GNNs.
Besides, to ensure that the proposed relation embeddings can acquire effective information, a gradient scaling factor is proposed to adjust the updates in each optimizing iteration, which constrains the weights of the relations to converge to a proper value.
A detailed illustration of our RE-GNN is presented in \cref{fig2-framework}.

\subsection{Feature Projection}

Usually, the features in different types of nodes are extracted via various schemes, while these nodes are always associated in certain conditions.
Therefore, the features in each node are firstly projected via a type-specific linear transformation \cite{han,gtn}.
The projected features are represented as
\begin{equation}
    \tilde{x}_{i} = x_{i}W_{\phi(i)},
\end{equation}
where $x_i$ is the original features in node $i$ and $W_{\phi{(i)}}$ represents the learnable projection matrix for the node type $\phi{(i)}$.
Let $\tilde{X}=(\tilde{x}_1,\tilde{x}_2,...,\tilde{x}_n)$ be the matrix form of the features after transformation.

\subsection{Relation Embeddings}
After projecting the attributes of each type of nodes into the same feature space, nodes can exchange message with all their neighbors from different types of relations.
In the aggregation step, the messages from different types of edges should possess different weights (importances).
Then, one relation embedding value is utilized for each edge type to learn its importance.
The learned weighted adjacency matrix is formulated as 
\begin{equation}
    A_{H}^{(l)} = \sum_{i\in\mathcal{R}} \tau(e_i^{(l)}) A_i, 
    \label{original-relation-embeddings}
\end{equation}
where $e_i^{(l)}\in\mathbb{R}$ stands for the learnable relation embedding of edge type $i\in \mathcal{R}$ for the $l$-th layer, and $\tau(\cdot)$ is a function to generate the aggregation importances from the embeddings.
For simplicity, we set $\tau(\cdot)$ to be the LeakyReLU function to ensure the importances of different relations to be non-negative.
In different layers, the relation embeddings can learn different importances for various relation types.

Similarly, we consider the self-loop connections, which connect the node with itself, as special types of relations.
Here, similar embedding values are employed to learn the importances of node-type-specific self-loop connections.
The adjacency matrix with self-loops is formulated as
\begin{equation}
    \hat{A}_{H}^{(l)} = A_{H}^{(l)} + \sum_{j\in\mathcal{F}} \tau(s_{j}^{(l)}) I_j,
    \label{original-selfloop-embeddings}
\end{equation}
where $I_j$ is the diagonal matrix and $(I_j)_{ii} = 1$, if and only if $\phi(i) = j$.

Then, the learned weighted adjacency matrix can be utilized by the homogeneous GNN layers, i.e.,
\begin{equation}
    H^{(l+1)} = GNNLayer(\hat{A}_H^{(l)}, H^{(l)}),
\end{equation}
where $H^{(0)}=\tilde{X}$ represents the projected node features and $GNNLayer(\cdot,\cdot)$ can be the layers in any homogeneous GNNs, such as GCN and GAT.
If GCN is employed as the baseline model, the RE-GCN layers can then be represented by
\begin{equation}
    H^{(l+1)} = \sigma(\tilde{A}_H^{(l)}H^{(l)}W^{(l)}),
\end{equation}
where $\tilde{A}_H^{(l)}$ is the normalized version of $\hat{A}_H^{(l)}$.
Note that the symmetric normalized version is $\tilde{A}_H = \hat{D}_{H}^{-\frac{1}{2}}\hat{A}_H\hat{D}_{H}^{-\frac{1}{2}}$, which is usually utilized in processing the undirected graphs. 
Meanwhile, the asymmetric normalized version is $\tilde{A}_H=\hat{D}^{-1}_H\hat{A}_H$, which is usually applied to the directed graphs.
Here, $\hat{D}_H$ denotes the degree diagonal matrix corresponding to $\hat{A}$.

\subsection{Gradient Scaling}
\label{sec-gradient-scaling}
In the optimization process, the relation embeddings are firstly initialized as ones, where each type of relations has identical importance.
As the model training progresses, the weights of the relations are expected to diversify.
Since the absolute values of the relation embeddings are much larger than the other parameters, the learned embeddings are not differentiable enough.
For example, a regular parameter $w$ is initialized via the Xavier Uniform Initialization\cite{glorot} method by setting $|w|<a$, where $a$ is a small value correlated to the input and output dimensions and the relation embeddings are set to ones.
Due to the above \textit{numerical inconsistency}, it is difficult for the popular optimizers to simultaneously optimize the relation embeddings and model parameters.
To tackle this problem, a gradient scaling factor is proposed to enlarge the gradients of the weights of the relations.

Specifically, a pre-defined scaling factor $\lambda > 0$ is exploited on the original embeddings as
\begin{equation}
    \alpha_i=\lambda e_i.
    \label{scale-gradient}
\end{equation}
Then, the scaled weights are utilized to generate the adjacency matrix as
\begin{equation}
    A_H = \sum_{i\in\mathcal{R}} \tau(\alpha_i) A_i.
    \label{eq-a-h}
\end{equation}
To initialize the scaled weights as ones, each embedding parameter $e_i$ is initially set to $\frac{1.0}{\lambda}$.
The value of the scaling factor $\lambda$ will affect the updating modification to $\alpha$.
In general, if $\lambda \geq 1$, \cref{scale-gradient} serves as a gradient enlarging machine and vice versa.
To better illustrate the proposed gradient scaling factor, here we reveal how this scaling factor $\lambda$ functions in the optimization process.

By denoting the loss function as $\mathcal L$, the gradient of each relation embedding $e_i$ in Eq. \eqref{original-relation-embeddings} is $g_i = \frac{\partial{\mathcal{L}}}{\partial{e_i}}$.
For the gradient based optimizers, the parameters are updated via $e_i^{next} = e_i - \Delta e_i$, where $\Delta e_i= \kappa(g_i)$.
By utilizing the scaling factor in Eq. \eqref{scale-gradient}, the gradient to the embedding parameter $e_i$ is $\lambda g_i$.
Then, the updating modification becomes $\kappa(\lambda g_i)$.
For the common gradient optimizers, such as SGD, Momentum \cite{momentum}, and Nesterov \cite{Nesterov}, $\kappa(\lambda g_i) = \lambda \kappa(g_i)$.
For the adaptive optimizers like Adagrad \cite{adagrad} and Adam \cite{adam}, $\kappa(\lambda g_i) = \kappa(g_i)$, due to their adaptive learning strategies.
By considering the modifications of the weights, since the scaling factor $\lambda > 0$, the modification $\Delta \alpha_i$ becomes $\lambda^2 \Delta\alpha_i$ for the common optimizers like SGD, and $\lambda\Delta\alpha_i$ for the adaptive optimizers like Adam.
The detailed proof can be found in Appendix \ref{appendix:scaling-factor}.

In general, with the proposed scaling factor $\lambda$, the weights of the relations can be optimized within a suitable numerical interval.
Similarly, the adjacency matrix with self-loops can be obtained via
\begin{equation}
    \hat{A}_H = \sum_{i\in\mathcal{R}}\tau(\alpha_i) A_i + \sum_{j\in\mathcal{F}}\tau(\beta_j) I_j,
    \label{eq-hat-a-h}
\end{equation}
where $\beta_j = \lambda s_j$.

\subsection{Memory and Computational Complexities}
Each RE-GNN layer only employs one parameter for each relation type to model the importance of the corresponding relation.
Generally, our RE-GNN only introduces $|\mathcal{F}|+|\mathcal{R}|$ more parameters than the corresponding homogeneous GNN in each layer.
For the computational complexity, each RE-GNN layer only introduces $O(|\mathcal{E}|+|\mathcal{V}|)$ more calculations to compute the weighted adjacency matrix by \cref{eq-hat-a-h}, which can usually be neglected by the computational complexity of the corresponding homogeneous GNN layer, such as \cite{gcn,gat,gatv2}.
Considering both the memory and computational complexities, our RE-GNN is indeed efficient.

\section{Analysis}

In this section, we analyze the effectiveness of our RE-GNN from two distinct perspectives. 
Firstly, we provide a spatial analysis which intuitively illustrates the effectiveness of our framework.
Then, we reveal that our RE-GNN can be interpreted as a meta-path based heterogeneous GNN.
Specifically, when employing GCN as the backbone, our RE-GCN can degenerate to Graph Transformer Network (GTN) \cite{gtn}, which is a powerful meta-path based heterogeneous GNN and can generate meta-paths adpatively.
At last, we theoretically demonstrate that our RE-GCN possesses more expressive power than GTN.

\subsection{Spatial Analysis}

Under the assumption that all types of nodes are implicitly correlated, our RE-GNN firstly projects the raw features in different types of nodes into the same feature space.
Then, RE-GNN learns a weighted relation adjacency matrix for the homogeneous GNN, to model the importance of different relations.
For example, for a paper node in the ACM network, different messages from different relations (e.g., Paper-Paper, Conference-Paper, Subject-Paper and node-type-specific self-loop (Paper)) are aggregated based on the corresponding relational importances. 
Similarly, for a conference, the message passed by Paper-Conference relation and self-loop (Conference) connection are aggregated based on the weights of different relations.
A detailed case study can be found in \cref{sec:case stduy}.

\subsection{Meta-Path based Analysis}

Besides of explaining RE-GNN from the perspective of spatial aggregations, RE-GNN can also be interpreted as a meta-path based heterogeneous GNN, implicitly.
For convenience, Graph Transformer Network (GTN) \cite{gtn} is selected as the compared heterogeneous GNN in this subsection.
Our RE-GNN framework also employ GCN \cite{gcn} as the backbone, similar to GTN. 
Here, we reveal that our RE-GCN can degenerate to GTN in certain conditions.
Besides, we also theoretically demonstrate that RE-GCN possesses more expressive power than GTN.

\subsubsection{Simplifying RE-GCN}

As stated in \cref{sec-gtn}, GTN can be regarded as an ensemble of GCNs on multiple generated composite relations. 
By excluding the ensemble trick in Eq. \eqref{eq-gtn-3}, we consider a simple case, where GTN only learns a 2-length composite relation, as below.
\begin{equation}
    Z_P = \sigma\left(\tilde{A}_2H^{(0)}W^{(0)}_P\right),
    \label{eq-2-length-gtn}
\end{equation}
where $\tilde{A}_2 = \hat{D}^{-1}_2\hat{A}_{2}$, $\hat{A}_2 = A_{2} + I$, and $\hat{D}_2$ represents the degree matrix of $\hat{A}_2$.
Note that $A_{2} = \left(\sum_{i\in\mathcal{R}} \alpha_{1,i} A_i\right) \left(\sum_{i\in\mathcal{R}} \alpha_{2,i} A_i\right)$ is the learned adjacency matrix of 2-length meta-path.
By stacking two RE-GCN layers, we can obtain 
\begin{equation}
    Z_H = \sigma\left(\tilde{A}^{(1)}_H\sigma\left(\tilde{A}^{(0)}_HH^{(0)}W^{(0)}_H\right)W^{(1)}_H\right),
    \label{eq-2-layered-REGCN-2}
\end{equation}
where $\tilde{A}^{(i)}_H = (\hat{D}_H^{(i)})^{-1}\hat{A}^{(i)}_H, i=0,1$, and $\hat{A}^{(i)}$ is calculated via Eq. \eqref{eq-hat-a-h}.
If we simplify Eq. \eqref{eq-2-layered-REGCN-2} by removing the nonlinearity and coarsening the weight matrices (like SGC \cite{sgc}), it will degenerate to
\begin{equation}
    Z_{H,SGC} = \sigma\left(\tilde{A}_{H,2}H^{(0)}W_H\right),
    \label{eq-2-layers-SGC}
\end{equation}
where $\tilde{A}_{H,2}=\hat{D}^{-1}_{H,2}\hat{A}_{H,2}$ and $\hat{A}_{H,2}=\hat{A}^{(1)}_H\hat{A}^{(0)}_H$.
Note that $\hat{D}_{H,2}=\hat{D}_H^{(1)}\hat{D}_H^{(0)}$ is the degree matrix of $\hat{A}_{H,2}$.
After simplifying 2 RE-GCN layers, we actually obtain a 2-length GTN layer.
Note that the only difference between RE-GCN and GTN is $\hat{A}_{H}$ (RE-GCN) and $\hat{A}_{P}$ (GTN), which are generated from different formulas.
$\hat{A}_{H}$ (RE-GCN) contains extra self-loop relation embeddings, while $\hat{A}_P$ (GTN) only adds an identity self-loop matrix. 
The above observation indicates that RE-GCN can implicitly learn the composite relations.

Since RE-SGC is a simplified version of RE-GCN, which will give superior expressive power than RE-SGC, 
we can intuitively conclude that the 2-layered RE-GCN in Eq. \eqref{eq-2-layered-REGCN-2} possesses more expressive powers than the 1-layered GTN in Eq. \eqref{eq-2-length-gtn}.
Then, we will prove it theoretically and extend it to the general case.

\subsubsection{Theoretical Analysis}

Due to limited computing and storing resources, the values of the input features and parameters are always finite for any neural network.
Usually, to ensure the stability and robustness of a neural network, the input and parameters are normalized to small values.
Under such circumstance, we firstly introduce the bounded set to quantitatively analyze the expressive power of neural network.
Note that all the proofs are provided in the appendices.

\noindent\textbf{Definition 1} (Bounded set).
\textit{
  A bounded set is a set $\mathcal D$ where $\forall d \in \mathcal D $, $|d|<k$.
  $k>0$ is a real number, which is the bound of $\mathcal D$.
  $\mathcal D$ can be a set of value, vector or matrix and $|\cdot|$ is a corresponding norm.
}

As stated above, we assume that the dynamic ranges of the network inputs and parameters are both bounded sets.
For example, a multilayer perceptron (MLP) layer is a function $f: \mathcal{H}_{in}\rightarrow \mathcal{H}_{out}$, which is defined on a bounded set $\mathcal{H}_{in}\subset \mathbb{R}^{d_{in}}$ and $f$'s range is also a bounded set $\mathcal{H}_{out}\subset \mathbb{R}^{d_{out}}$.
$\forall h_{in} \in \mathcal{H}_{in}$ and $\forall h_{out} \in \mathcal{H}_{out}$, we can obtain $||h_{in}||_2^2 < k_{in}$ and $||h_{out}||_2^2 < k_{out}$, respectively.
$\forall h_{in}\in \mathcal{H}_{in}$, $f$ is defined as 
\begin{equation}
  f(h_{in}) = \sigma(h_{in}W+b),
\end{equation}
where $W\in \mathcal{W}\subset \mathbb{R}^{d_{in}\times d_{out}}$ is the learnable weight matrix and $\sigma(\cdot)$ is a ReLU function.
The bound of $W$ is defined as $k_w=|W|\overset{\text{def}}{=}\max ||w_j||_2^2$, where $w_j$ is the column vector of $W$.
Note that $b$ is a bias vector and its bound is defined as $k_b=\max|b_i|$.

\noindent\textbf{Definition 2} (Bound of Multilayer Perceptron Layer).
\textit{
  A multilayer perceptron (MLP) layer $f(W, b): \mathcal{H}_{in}\rightarrow \mathcal{H}_{out}$ is defined on a bounded set $\mathcal{H}_{in}\subset \mathbb{R}^{d_{in}}$.
  $k_{in}, k_w, k_b$ are the bounds of the input $h_{in}$, parameter $W$ and bias $b$, respectively.
  The bound of $f$ is defined as $k=\max(k_{in}, k_w, k_b)$.
}

Since it is usually necessary to constrain the bound of input data when training or designing a neural network, 
we take the bound of the input into account for an MLP layer.
According to the definition above, $k$ is also the bound of the input, parameters, and bias.
Subsequently, we can infer the rationality of an MLP layer based on its bound.
Here, we quantitatively analyze the expressive power of a composite of two MLP layers, compared to that of only one MLP layer.

\noindent\textbf{Lemma 3.}

\textit{
  Given an MLP layer $f: \mathcal{H}_{in}\rightarrow \mathcal{H}_{out}$ bounded by $k$, there exists a composite of two MLP layers $(f_1\circ f_2)$ which equals to $f$, where $f_1$ is also bounded by $k$ and $f_2$ is bounded by $\max\{1, 2k\}$.
}

Since a particular solution of the layer $f_2$ is $W_2$ being an identity matrix when $b_2$ is bounded by $2k$, the layer $f_2$ is bounded by $\max(1, 2k)$ in Lemma 3.
By considering that the bias plays a complementary role to parameter $W$, we do not separately restrict the bound of $b$, in practice.
Thus, it is acceptable that the bias of $f_2$, i.e., $b_2$, has a bit larger bound. 

According  to Lemma 3, we can conclude that a composite of two MLP layers possesses no less expressive power than one MLP layer. 

Now, let us consider the scenario of GNN.
Similarly, for a GCN layer, we still utilize $k=\max(k_{in}, k_w, k_b)$ as its bound, where $k_{in}$, $k_w$ and $k_b$ are the bounds of the input, parameter and bias, respectively.
In practice, GCN layer usually utilizes a bias vector.
For example, the GTN layer in Eq. (13) can be rewritten as
\begin{equation}
  Z_{P} = \sigma\left(\tilde{A}_2 H W_P + B_P\right),
  \label{gtn-B}
\end{equation}
where $B=
\begin{pmatrix}
  b\\
  \vdots\\
  b
  \end{pmatrix}$
is the \textit{broadcast} of a bias vector $b$ (stacking multiple rows with a row vector $b$).
The GTN layer actually adds a bias vector to each sample.
Similarly, a composite of two RE-GCN layers in Eq. (14) can be rewritten as
\begin{equation}
  Z_H = \sigma\left(\tilde{A}^{(1)}_H\sigma\left(\tilde{A}^{(0)}_HXW^{(0)}_H+B^{(0)}_H\right)W^{(1)}_H+B^{(1)}_H\right).
\end{equation}

\noindent\textbf{Lemma 4.}

\textit{
  If a vector set $\mathcal{H}$ is bounded by $k$.
  Denote $o=p_1h_1+p_2h_2+...+p_rh_r$, where $\sum_{i=1}^r p_i=1$, $h_i\in\mathcal{H}, i=1,2,...,r$.
  Then, $o$ is also bounded by $k$, i.e., $||o||_2^2 < k$.
}

Then, according to Lemma 3 and 4, Corollary 5 can be obtained.

\noindent\textbf{Corollary 5.}

\textit{
  If $f_{GTN}: \mathcal{G}\rightarrow \mathbb{R}^{N\times C}$ is a 2-lengthed GTN layer, which is bounded by $\xi$,
  there exists a composite of two RE-GCN layers, where the first layer is bounded by $\xi$ and the second layer is bounded by $\max\{1, 2\xi\}$, which is equivalent to $f_{GTN}$.
}

Corollary 5 theoretically reveals that a composite of two RE-GCN layers possesses no less expressive powers than a 2-lengthed GTN layer.
Then, we can extend it to the general case.

\noindent\textbf{Theorem 6.}

\textit{
    Let $G\in\mathcal{G}$ be a heterogeneous graph, where the node features $X\in\mathcal{X}$ are normalized, i.e., $\forall X\in\mathcal{X}, |X|<\xi$.
    If a non-ensembled GTN, $m: \mathcal{G}\rightarrow \mathbb{R}^{N\times C}$, maps the nodes in $G$ to any node embeddings $Z\in\mathbb{R}^{N\times C}$, there exists a RE-GCN which is equivalent to GTN.
}

Theorem 6 further demonstrates that RE-GCN possesses no less expressive power than the GTN. 
Then, Theorem 7 is obtained as follows.

\noindent\textbf{Theorem 7.}

\textit{
    Let $G\in\mathcal{G}$ be a heterogeneous graph, where the node features $X\in\mathcal{X}$ are normalized, i.e., $\forall X\in\mathcal{X}, |X|<\xi$.
    There exists a RE-GCN, $r: \mathcal{G}\rightarrow \mathbb{R}^{N\times C}$, which can map $G$ to the node embeddings $Z\in\mathbb{R}^{N\times C}$ that GTN cannot map $G$ to.
}

Theorems 6 and 7 jointly prove that our RE-GCN possesses more expressive power than the GTN.
According to the above analysis, we can conclude that RE-GCN can be interpreted as an implicit meta-path based heterogeneous GNN, which indicates that it may be unnecessary to design or generate the meta-path explicitly.
By simply stacking multiple RE-GNN layers, the heterogeneity may also be efficiently handled.

\begin{table}[t]
  \small
  \centering
  \caption{Statistics of the datasets.}
  \begin{tabular}{|l|l|l|}
  \hline
  Datasets & Nodes                                                                                                                                        & Edges                                                                                                                   \\
  \hline
  DBLP     & \begin{tabular}[c]{@{}l@{}}\# author(A): 4,057\\ \# paper(P): 14,328\\ \# term(T): 7,723\\ \# venue(V): 20\end{tabular}                      & \begin{tabular}[c]{@{}l@{}}\# A-P:19,645\\ \# P-T:85,810\\ \# P-V:14,328\end{tabular}                                   \\ 
  \hline
  ACM      & \begin{tabular}[c]{@{}l@{}}\# paper(P): 4,019\\ \# author(A): 7,167\\ \# subject(S): 60\end{tabular}                                         & \begin{tabular}[c]{@{}l@{}}\# P-P:9,615\\ \# P-A:13,407\\ \# P-S:4,019\end{tabular}                                     \\ 
  \hline
  IMDB     & \begin{tabular}[c]{@{}l@{}}\# movie(M): 4,278\\ \# director(D): 2,081\\ \# actor(A): 5,257\end{tabular}                                      & \begin{tabular}[c]{@{}l@{}}\# M-D:4,278\\ \# M-A:12,828\end{tabular}                                                    \\ 
  \hline
  OGBN-MAG & \begin{tabular}[c]{@{}l@{}}\# paper (P): 736,389\\ \# author (A): 1,134,649\\ \# institutions (I): 8,740\\ \# fields(F): 59,965\end{tabular} & \begin{tabular}[c]{@{}l@{}}\# A-I: 1,043,998\\ \# A-P: 7,145,660\\ \# P-P: 7,505,078 \\ \# P-F: 10,792,672\end{tabular} \\ 
  \hline
  \end{tabular}
  \label{table-datasets}
\end{table}

\begin{table*}[t]
  \centering
  \caption{Performances on the node classification task.}
  \begin{tabular}{c|cc|cc|cc}
      \toprule
      \multirow{2}{*}{\textbf{Methods}} & \multicolumn{2}{c|}{\textbf{DBLP}} & \multicolumn{2}{c|}{\textbf{ACM}}                & \multicolumn{2}{c}{\textbf{IMDB}} \\
                    & Macro-F1               & Micro-F1               & Macro-F1               & Micro-F1               & Macro-F1               & Micro-F1        \\
      \midrule
      Metapath2vec  & 91.71 ± 0.00           & 92.39 ± 0.00           & 75.34 ± 0.00           & 76.79 ± 0.00           & 48.70 ± 0.00           & 50.40 ± 0.00           \\
      \midrule
      HAN           & 92.45 ± 0.87           & 92.96 ± 0.83           & 91.80 ± 0.38           & 91.67 ± 0.31           & 57.69 ± 0.91           & 58.15 ± 0.93           \\
      GTN           & 93.55 ± 0.22           & 94.29 ± 0.21           & 91.92 ± 0.67           & 91.81 ± 0.65           & 58.63 ± 1.45           & 60.19 ± 1.66           \\
      MAGNN         & 93.39 ± 0.33           & 93.89 ± 0.30           & 91.49 ± 1.18           & 91.46 ± 1.08           & 58.75 ± 2.04           & 59.48 ± 1.45           \\
      MAGNN-AC      & 93.00 ± 0.29           & 93.58 ± 0.21           & 88.34 ± 3.55           & 88.60 ± 2.98           & 56.98 ± 0.38           & 57.80 ± 0.44           \\
      \midrule
      R-GCN          & 92.26 ± 0.47           & 92.87 ± 0.42           & 93.36 ± 0.21           & 93.36 ± 0.21           & 58.96 ± 0.29           & 59.34 ± 0.27           \\
      HGT           & 91.43 ± 1.15           & 92.31 ± 0.91           & 92.40 ± 0.60           & 92.30 ± 0.54           & 58.16 ± 0.84           & 58.37 ± 0.70           \\
      HGB           & 93.20 ± 0.33           & 93.69 ± 0.29           & 93.15 ± 0.30           & 93.13 ± 0.27           & 58.11 ± 0.86           & 58.42 ± 0.71           \\
      R-HGNN        & 92.89 ± 0.59           & 93.43 ± 0.53           & 92.16 ± 0.71           & 92.10 ± 0.67           & 56.79 ± 1.01           & 57.27 ± 0.95           \\
      \midrule
      GAT           & 90.19 ± 0.89           & 90.94 ± 0.77           & 93.16 ± 0.16           & 93.16 ± 0.16           & 57.62 ± 1.20           & 58.92 ± 0.91           \\
      GCN           & 87.39 ± 0.23           & 88.31 ± 0.19           & 93.60 ± 0.29           & 93.58 ± 0.27           & 58.37 ± 1.26           & 59.59 ± 1.10           \\
      \midrule
      GAT-M         & 90.57 ± 0.70           & 91.15 ± 0.63           & 89.98 ± 0.52            & 89.94 ± 0.47           & 54.20 ± 0.84           & 54.80 ± 0.59           \\
      GCN-M         & 89.04 ± 0.71           & 89.99 ± 0.64           & 90.20 ± 0.24           & 90.15 ± 0.23           & 53.99 ± 0.95           & 54.72 ± 0.61           \\
      \midrule
      RE-GAT        & \textbf{95.06 ± 0.16}  & \textbf{95.41 ± 0.15}  & \textbf{94.04 ± 0.23}  & \textbf{93.99 ± 0.22}  & \textbf{60.01 ± 1.09}  & \textbf{60.53 ± 0.85}  \\
      RE-GCN        & \textbf{95.46 ± 0.29}  & \textbf{95.80 ± 0.27}  & \textbf{93.95 ± 0.16}  & \textbf{93.93 ± 0.16}  & \textbf{60.88 ± 0.95}  & \textbf{61.51 ± 0.64}  \\
      \bottomrule
  \end{tabular}
  \label{table-main-res}
\end{table*}

\begin{table*}[t]
  \centering
  \caption{Performances on the node clustering task.}
  \begin{tabular}{c|cc|cc|cc}
      \toprule
      \multirow{2}{*}{\textbf{Methods}} & \multicolumn{2}{c|}{\textbf{DBLP}} & \multicolumn{2}{c|}{\textbf{ACM}}                & \multicolumn{2}{c}{\textbf{IMDB}} \\
                    & NMI                   & ARI                   & NMI                   & ARI                   & NMI                   & ARI                 \\
      \midrule
      metapath2vec  & 78.61 ± 0.04          & 83.64 ± 0.04          & 41.80 ± 0.01          & 34.71 ± 0.01          & 5.49 ± 0.24           & 5.16 ± 0.27 \\
      \midrule
      HAN           & 76.30 ± 0.68          & 82.12 ± 0.56          & 70.71 ± 0.91          & 75.16 ± 0.88          & 12.96 ± 1.09          & 14.34 ± 1.28    \\
      MAGNN         & 79.89 ± 0.85          & 84.94 ± 0.77          & 70.70 ± 1.08          & 74.34 ± 1.42          & 15.70 ± 0.77          & \textbf{15.73 ± 1.19}           \\
      R-GCN         & 76.85 ± 1.70          & 82.73 ± 1.70          & 75.00 ± 0.80          & 77.82 ± 0.80          & 13.63 ± 0.16          & 15.59 ± 0.16    \\
      HGB           & 77.09 ± 1.93          & 82.86 ± 1.55          & 75.91 ± 0.84          & 80.56 ± 0.82          & 11.32 ± 2.48          & 11.40 ± 4.09    \\
      R-HGNN        & 75.16 ± 3.53          & 79.14 ± 6.78          & 63.86 ± 0.04          & 63.42 ± 0.08          & 9.67 ± 0.91           & 10.03 ± 1.72     \\
      \midrule
      GCN           & 60.59 ± 0.02          & 63.80 ± 0.03          & 77.28 ± 0.42          & 80.82 ± 0.56          & 14.15 ± 2.23          & 13.01 ± 2.43           \\
      RE-GCN        & \textbf{83.57 ± 1.78} & \textbf{88.13 ± 1.88} & \textbf{77.49 ± 0.58} & \textbf{81.67 ± 0.56} & \textbf{15.83 ± 1.10} & 15.65 ± 2.28  \\
      \bottomrule
  \end{tabular}
  \label{table-clustering}
\end{table*}

\section{Evaluations}

\subsection{Datasets}

\label{sec-datasets}

Four widely utilized heterogeneous datasets, including three heterogeneous academic networks (i.e., DBLP, ACM, and OGBN-MAG) and one heterogeneous movie network (i.e., IMDB), are employed to demonstrate the effectiveness of our RE-GNNs.
Their details are shown in \cref{table-datasets}.

In DBLP, four types of nodes, i.e., papers, authors, venues, and terms, and three types of edges (relations), i.e., A-P, P-T, and P-V, are constructed.
According to the conferences they are submitted to, authors are categorized into four research areas, i.e., Database, Data Mining, Artificial Intelligence and Information Retrieval.
Each paper is described by a bag-of-words (BOW) representation of its keywords, and each author is described by the BOW embeddings of its published papers.
Each term is represented by the Glove word vectors \cite{Glove}, and the attributes of venues are formed as one-hot vectors. 

In ACM, three types of nodes, i.e., papers, authors, and subjects, and three types of directed edges (relations) are constructed.
According to the conference they published, papers are classified into three categories, i.e., Database, Wireless Communication, and Data Mining.
Each paper is described by a BOW representation of keywords.
The attributes of authors and subjects are formed as one-hot vectors.

In IMDB, three types of nodes, i.e., movies, directors, and actors, and two types of directed edges (relations) are formed.
According to their genre, the movies are classified into three categories, i.e., Action, Comedy, and Drama.
Movies are described by the BOW embeddings of their plots.
The attributes of actors and directors are represented by the averaged attributes of their related movies.

In OGBN-MAG, four types of nodes, i.e., papers, authors, institutions, and fields of study, and four types of directed edges (relations), i.e., A-I, A-P, P-P, P-F, are constructed.
According to the venue (conference or journal) they published, papers are divided into 349 categories.
Each paper is associated with a 128-dimensional word2vec feature vector.
We employ metapath2vec \cite{metapath2vec} to generate the features in other types of nodes.

We process the relations with directions for all the datasets and consider their reversed ones.
For DBLP, ACM, and IMDB, we adopt the same data division strategies as \cite{magnn,hgnn-ac}, where 400/400 nodes are randomly selected as training/validation sets, and the remaining nodes (approximately 80\%) are selected as the test set.
For OGBN-MAG, we adopt its official splits \cite{ogb}.

\subsection{Baselines}
We compare our method with various kinds of popular GNN methods, including five homogeneous GNNs (i.e., GCN \cite{gcn} and GAT \cite{gat}, GIN \cite{gin}, GATv2 \cite{gatv2}, GraphSAGE \cite{graphsage}, and MixHop \cite{mixhop}), one heterogeneous graph embedding method (i.e., Metapath2vec \cite{metapath2vec}), four meta-path based heterogeneous GNNs (i.e., HAN \cite{han}, GTN \cite{gtn}, MAGNN \cite{magnn}, and MAGNN-AC \cite{hgnn-ac}), and four relation based heterogeneous GNNs (i.e., R-GCN \cite{rgcn}, HGT \cite{hgt}, HGB \cite{hgb}, and R-HGNN \cite{r-hgnn}).
Note that the homogeneous GNNs, such as GCN and GAT, are constructed on the homogeneous graph, where the heterogeneity is directly eliminated.
Here, GCN-M and GAT-M respectively represent the GCN and GAT models which are constructed on several meta-path based homogeneous graphs, and the best scores are reported.

\subsection{Implementation Details}

For the homogeneous GNNs, four-layered GNNs with 64 hidden neurons and 4 heads (only for GAT and GATv2), are employed as our baselines.
Note that we employ GAT and GATv2 with 128 hidden neurons and 1 heads on IMDB, which are obtained via hyperparameter search.
For the corresponding RE-GNNs, the gradient scaling factor $\lambda$ is set to 100.
In the training process, both the homogeneous GNNs and RE-GNNs are trained for a maximum of 200 epochs with an early stopping condition at 50 epochs.
The cross-entropy loss is utilized as the loss function.
Adam \cite{adam} optimizer is employed with the learning rate of 0.001, the weight decay rate of 0.001 for ACM and DBLP, and 0.005 for IMDB. 
The dropout layers are utilized for the input of each GNN layer in the training process with a dropout rate of 0.6.

\subsection{Node Classification}

\subsubsection{Comparisons}
The node classification results of three heterogeneous datasets, i.e., DBLP, ACM, and IMDB, are presented in \cref{table-main-res}.
For the employed baselines, both the meta-path based and relation based heterogeneous GNNs can achieve decent performances.
Besides, though homogeneous GNNs, i.e., GCN and GAT, can perform well on ACM and IMDB datasets, they are not suitable for DBLP, compared to other heterogeneous GNNs.
On the contrary, our RE-GNNs, which only utilize one-dimensional embeddings, can allow the homogeneous GNNs to process the heterogeneous graphs effectively.
For example, in DBLP, RE-GCN performs much better than the original GCN, which has 8.07 and 7.49 improvements in Macro-F1 and Micro-F1 scores, respectively.
A similar trend happens on GAT and RE-GAT.
Meanwhile, for the ACM dataset, though our RE-GCN still achieves the best performance, the performance gain is relatively small, e.g., 0.35 in the Macro-F1 score.
The reason may be that the weights of the relations with the current initialization are already close to the optimal weights in some particular cases. 
Under such circumstances, the performances of homogeneous GNNs may be close to or even approximately equal to the corresponding RE-GNNs.
Generally, our RE-GCN and RE-GAT outperform the other well-designed heterogeneous GNNs, which usually possess complicated modules.

\begin{figure}[t]
  \centering
  \subfloat[] {
    \label{fig-gnn:dblp}
    \includegraphics[width=0.48\columnwidth]{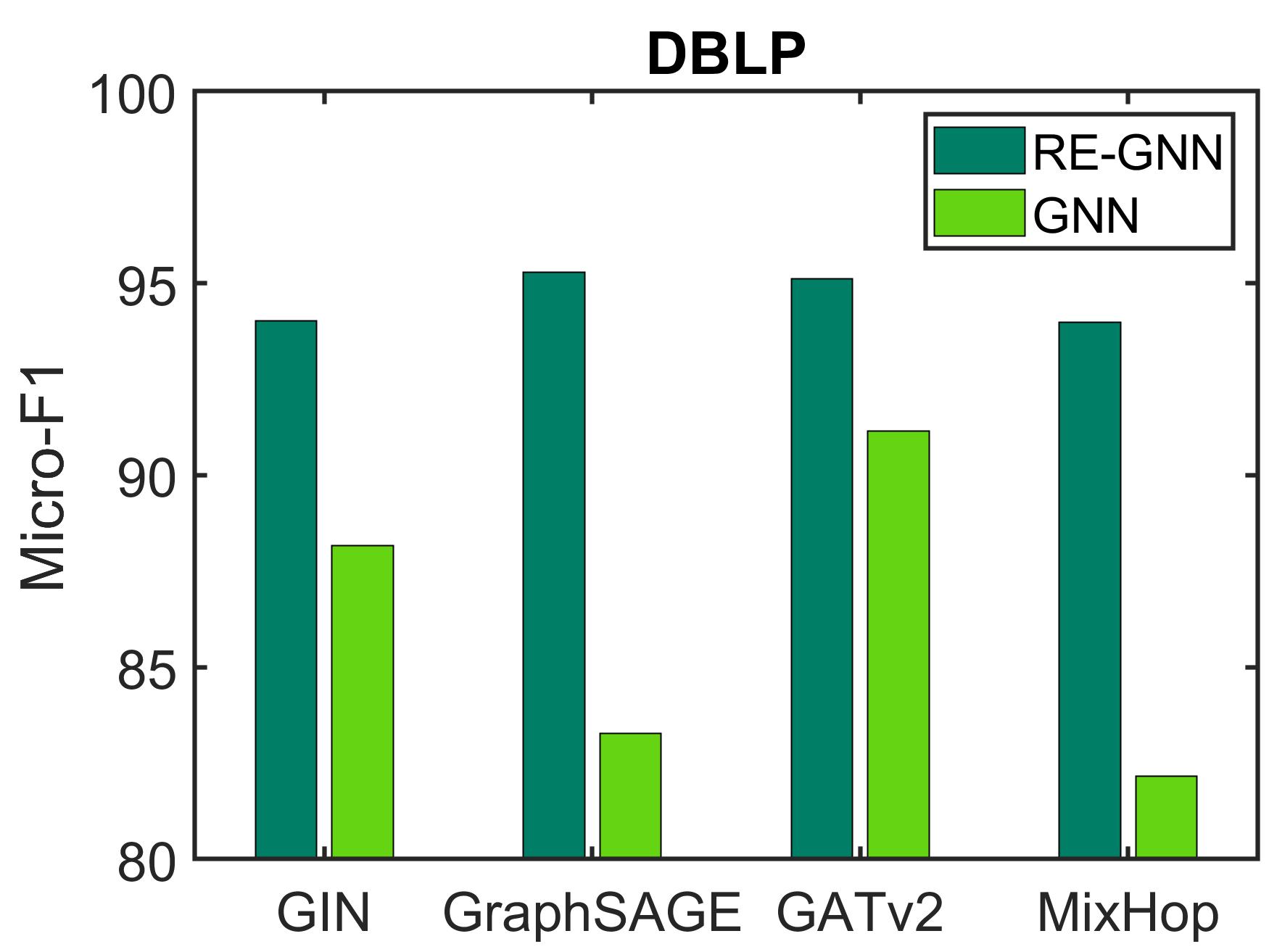}
  }
  \subfloat[] {
    \label{fig-gnn:acm}
    \includegraphics[width=0.48\columnwidth]{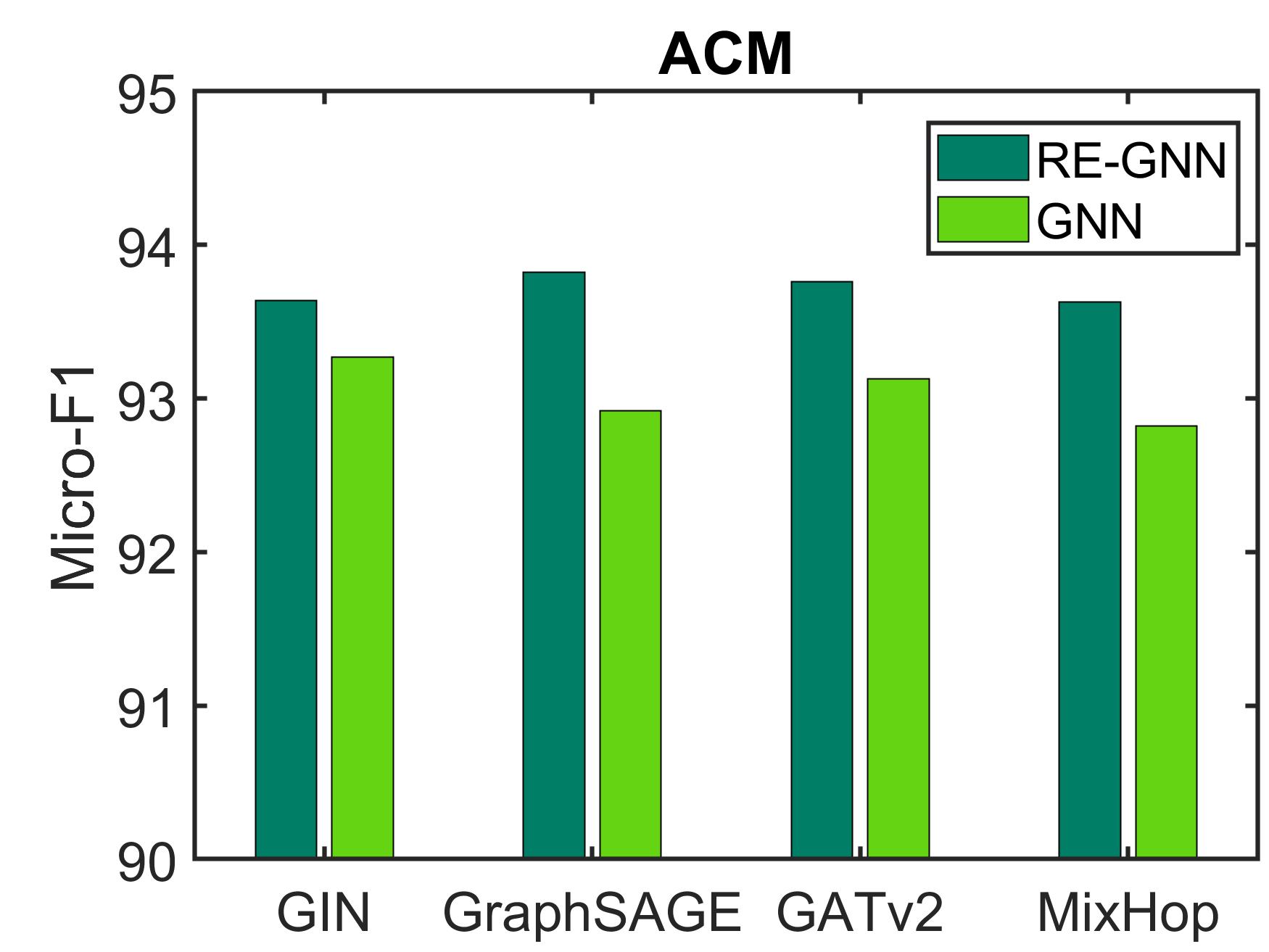}
  }
  \caption{Results of our RE-GNNs with other homogeneous GNNs as backbones.}
  \label{fig-gnn}
\end{figure}

\subsubsection{Applied to other GNNs}
The proposed framework can be easily applied to other homogeneous GNNs.
Besides of GCN and GAT, four other typical homogeneous GNNs, i.e., GIN \cite{gin}, GraphSAGE \cite{graphsage}, GATv2 \cite{gatv2}, and MixHop \cite{mixhop}, are employed for further validations.
As can be observed from \cref{fig-gnn}, though these homogeneous GNNs cannot perform well on DBLP, their corresponding RE-GNNs can achieve superior yet decent results.
On the ACM dataset, these homogeneous GNNs can achieve good results, similar to GCN and GAT.
By introducing our relation embeddings, the performances of the corresponding RE-GNNs also possess certain improvements.
In general, these results demonstrate that our RE-GNNs can effectively assign adequate abilities to the homogeneous GNNs to handle heterogeneous graphs.

\begin{figure*}[t]
  \centering
  \subfloat[Metapath2vec] {
    \label{fig-v:metapath2vec}
    \includegraphics[width=0.44\columnwidth,trim=50 50 50 50,clip]{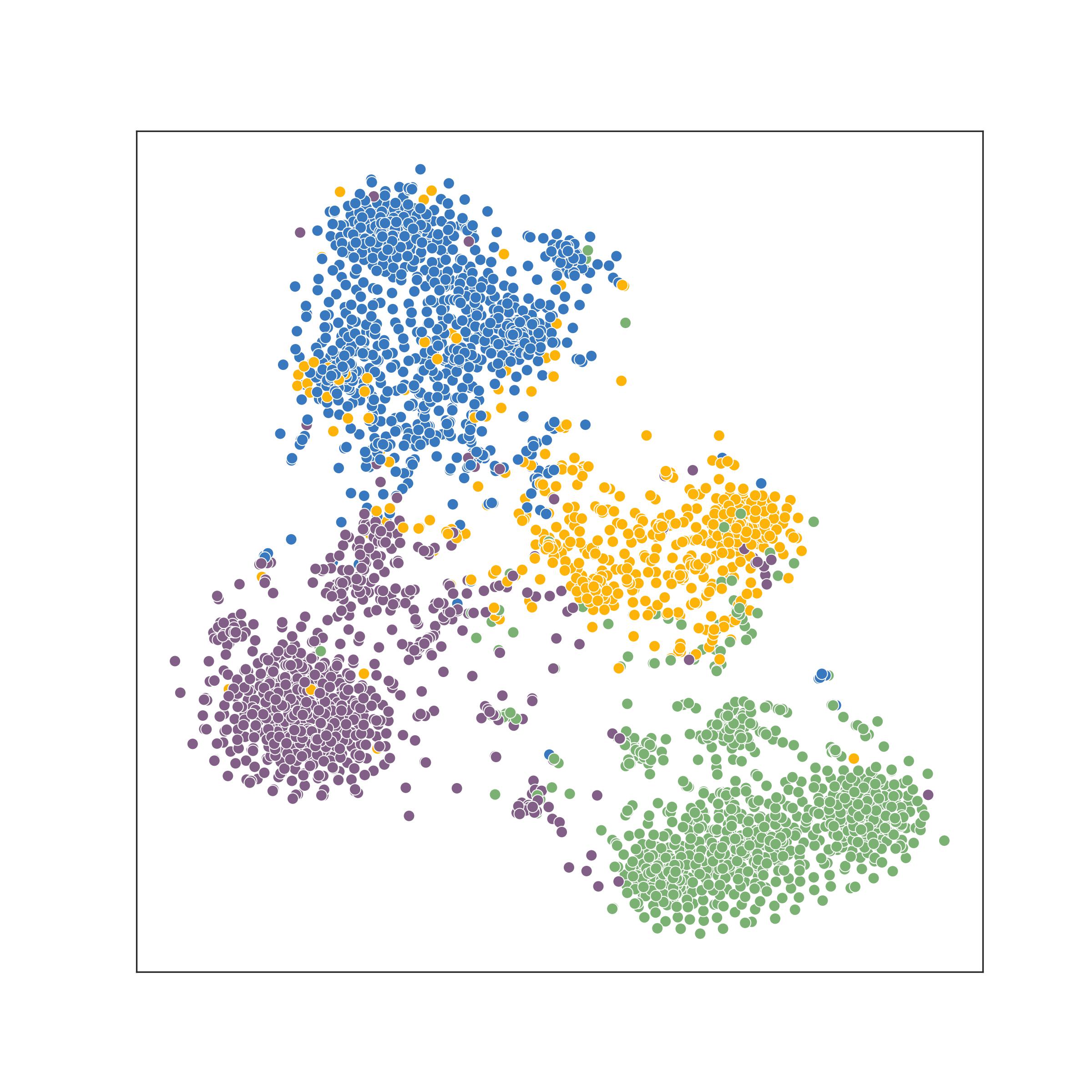}
  }
  \subfloat[HAN] { 
    \label{fig-v:han}
    \includegraphics[width=0.44\columnwidth,trim=50 50 50 50,clip]{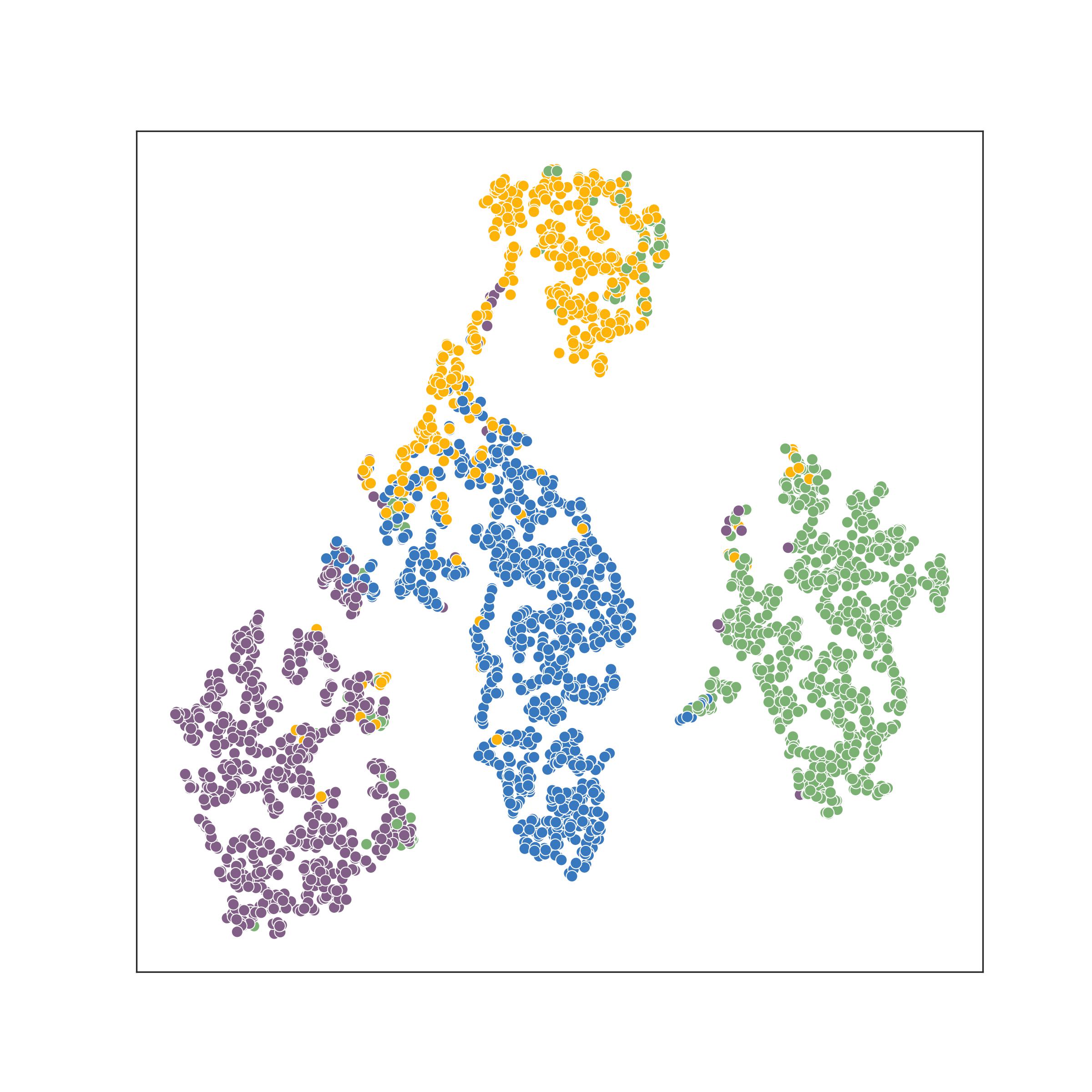}
  }
  \subfloat[MAGNN] { 
    \label{fig-v:magnn}
    \includegraphics[width=0.44\columnwidth,trim=50 50 50 50,clip]{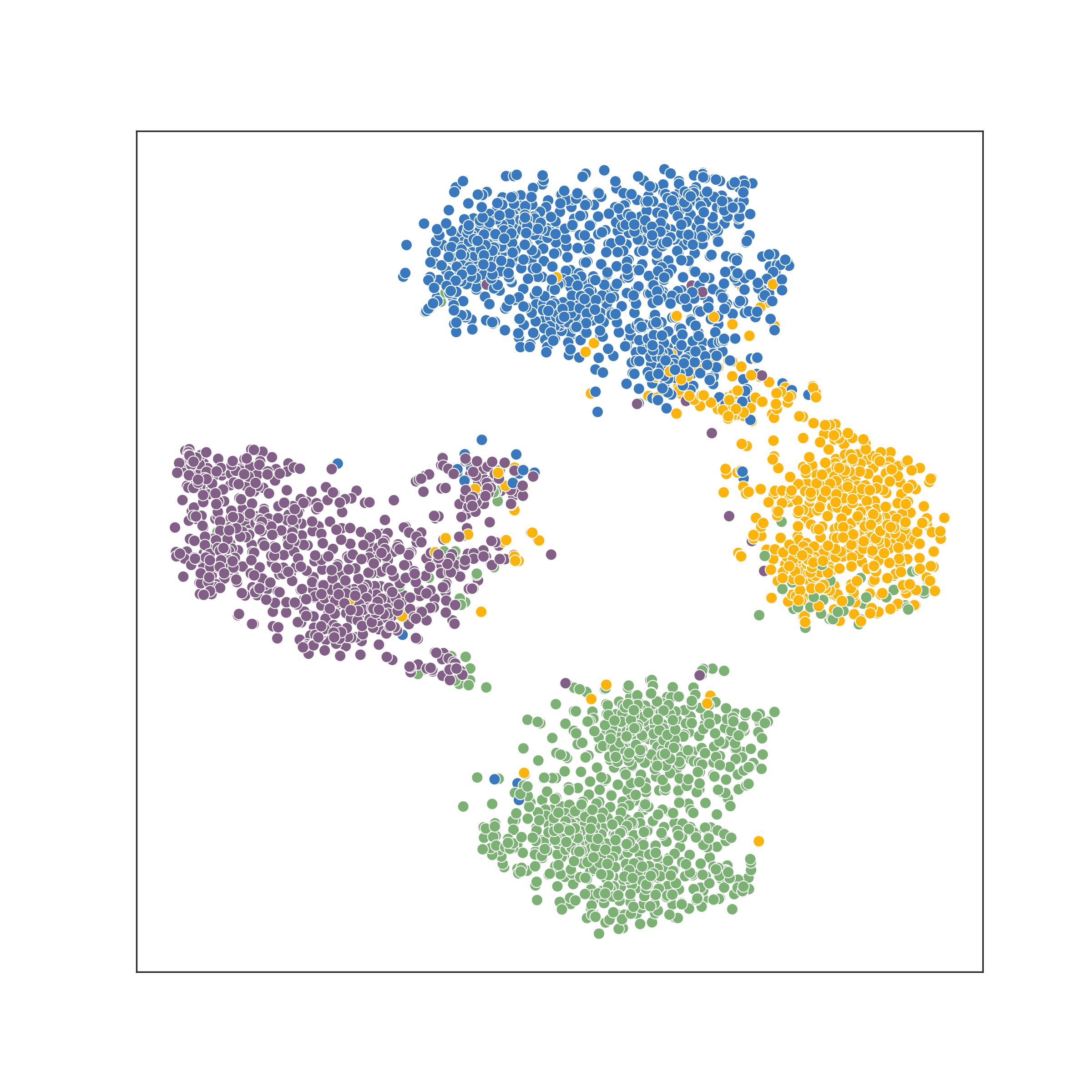}
  }
  \subfloat[R-GCN] { 
    \label{fig-v:gtn}
    \includegraphics[width=0.44\columnwidth,trim=50 50 50 50,clip]{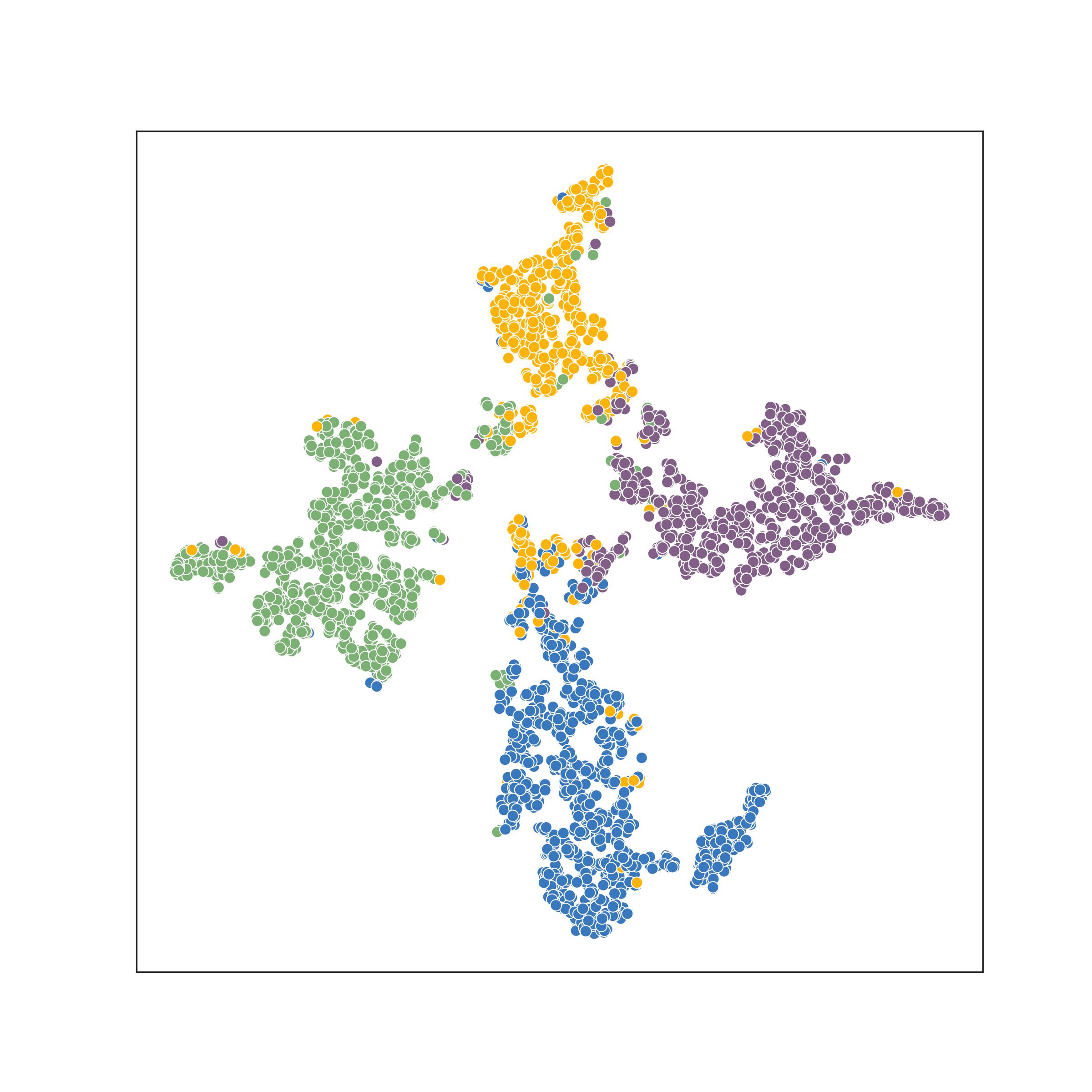}
  }

  \subfloat[HGB] { 
    \label{fig-v:hgb}
    \includegraphics[width=0.44\columnwidth,trim=50 50 50 50,clip]{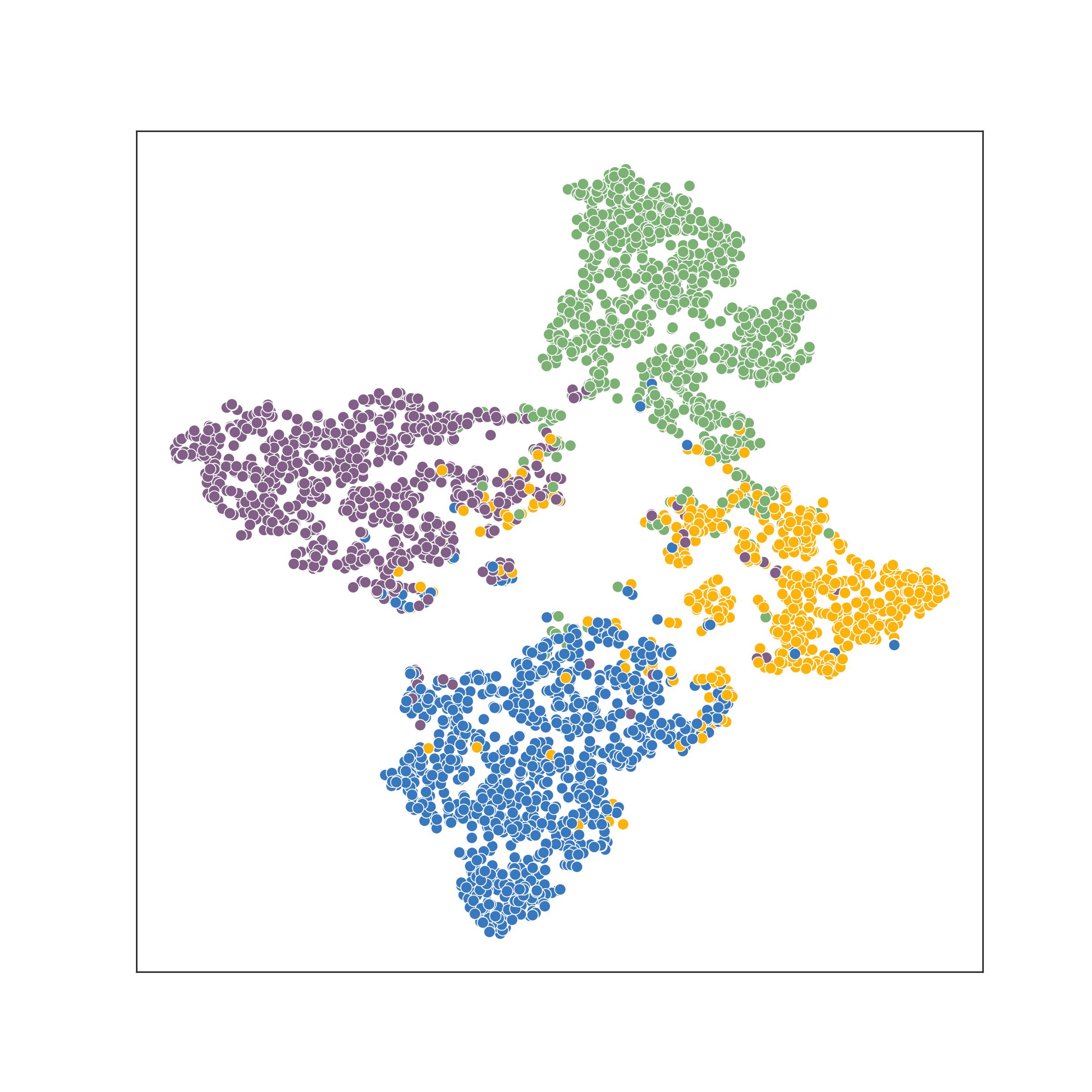}
  }
  \subfloat[RHGNN] { 
    \label{fig-v:rghnn}
    \includegraphics[width=0.44\columnwidth,trim=50 50 50 50,clip]{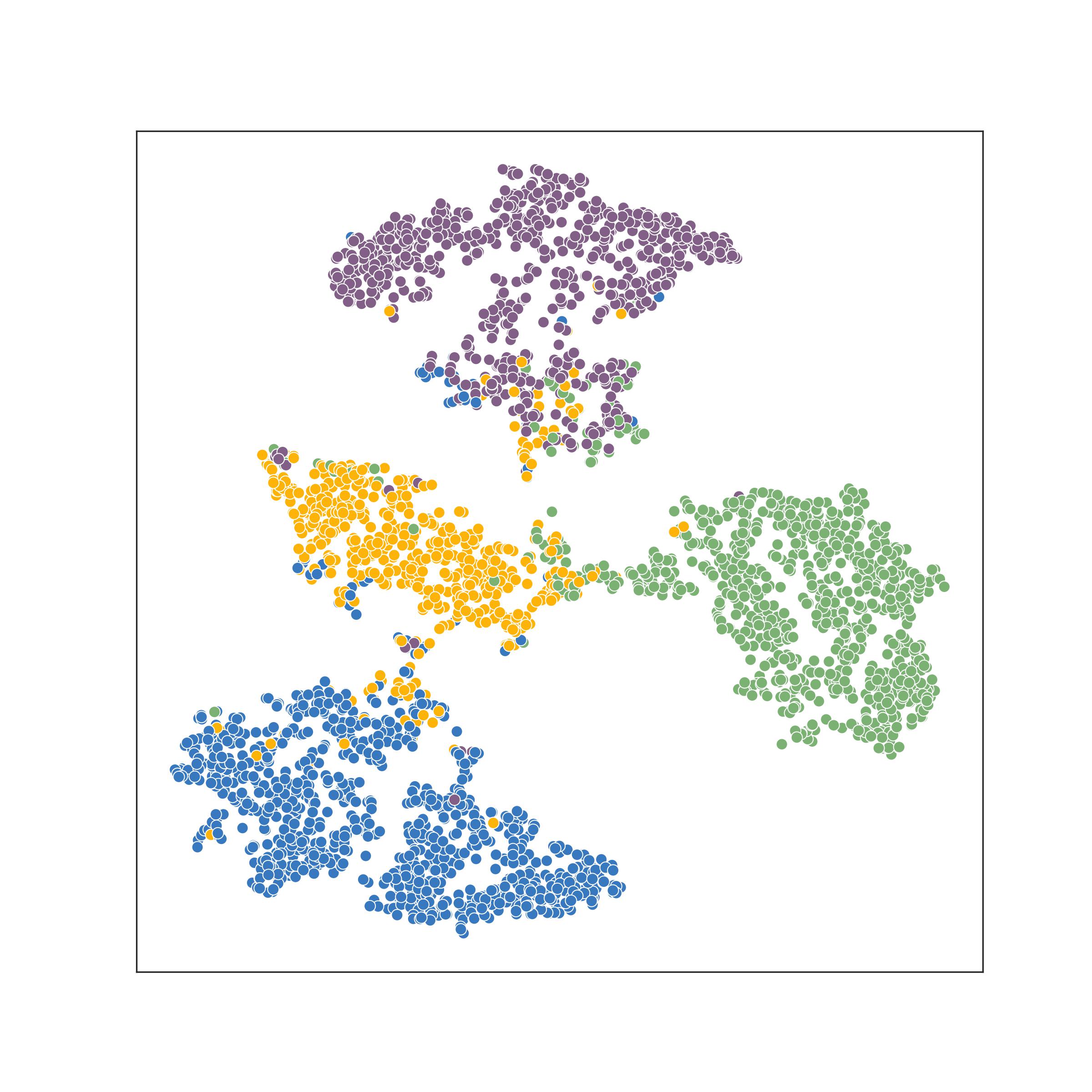}
  }
  \subfloat[GCN] { 
    \label{fig-v:gcn}
    \includegraphics[width=0.44\columnwidth,trim=50 50 50 50,clip]{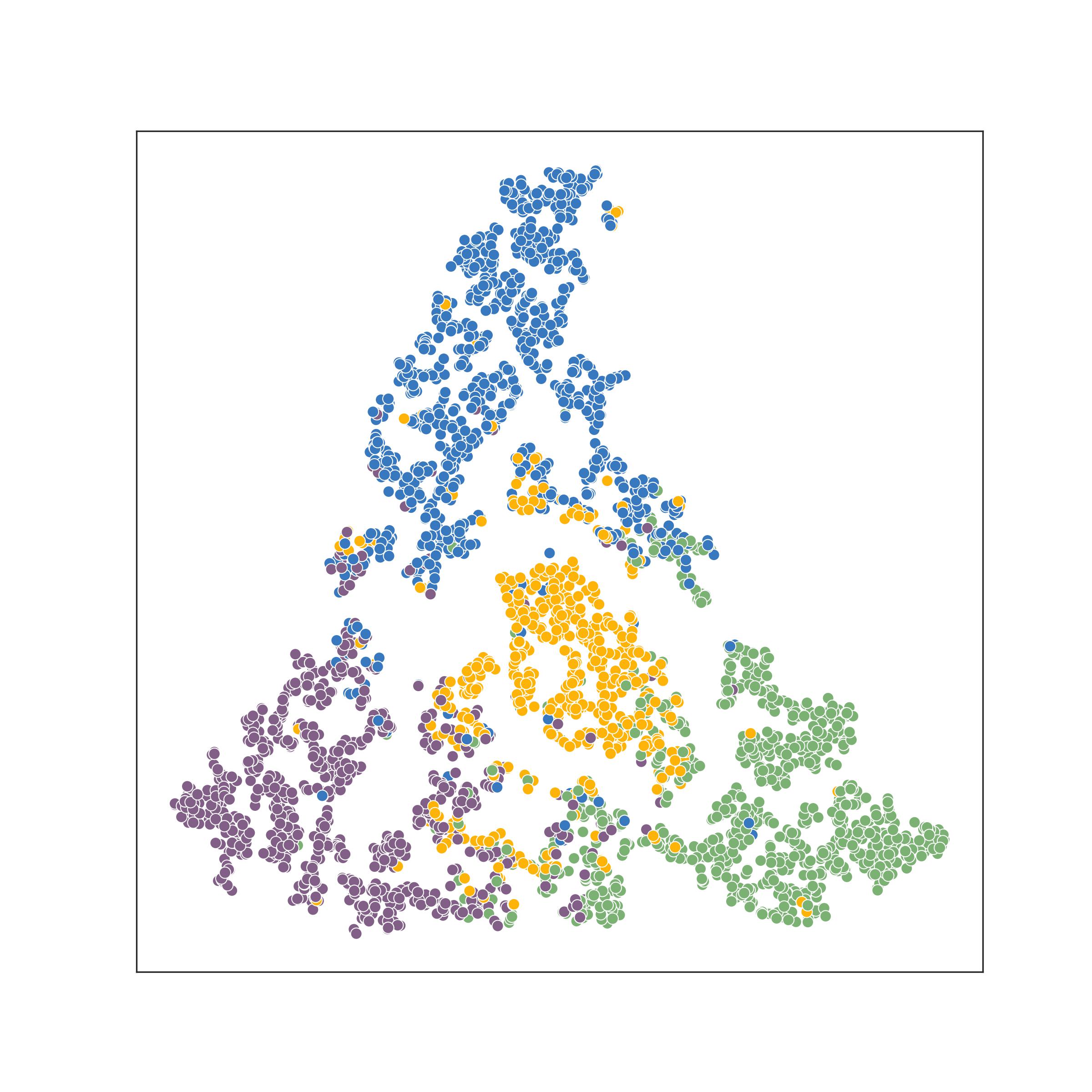}
  }
  \subfloat[RE-GCN] { 
    \label{fig-v:regcn}
    \includegraphics[width=0.44\columnwidth,trim=50 50 50 50,clip]{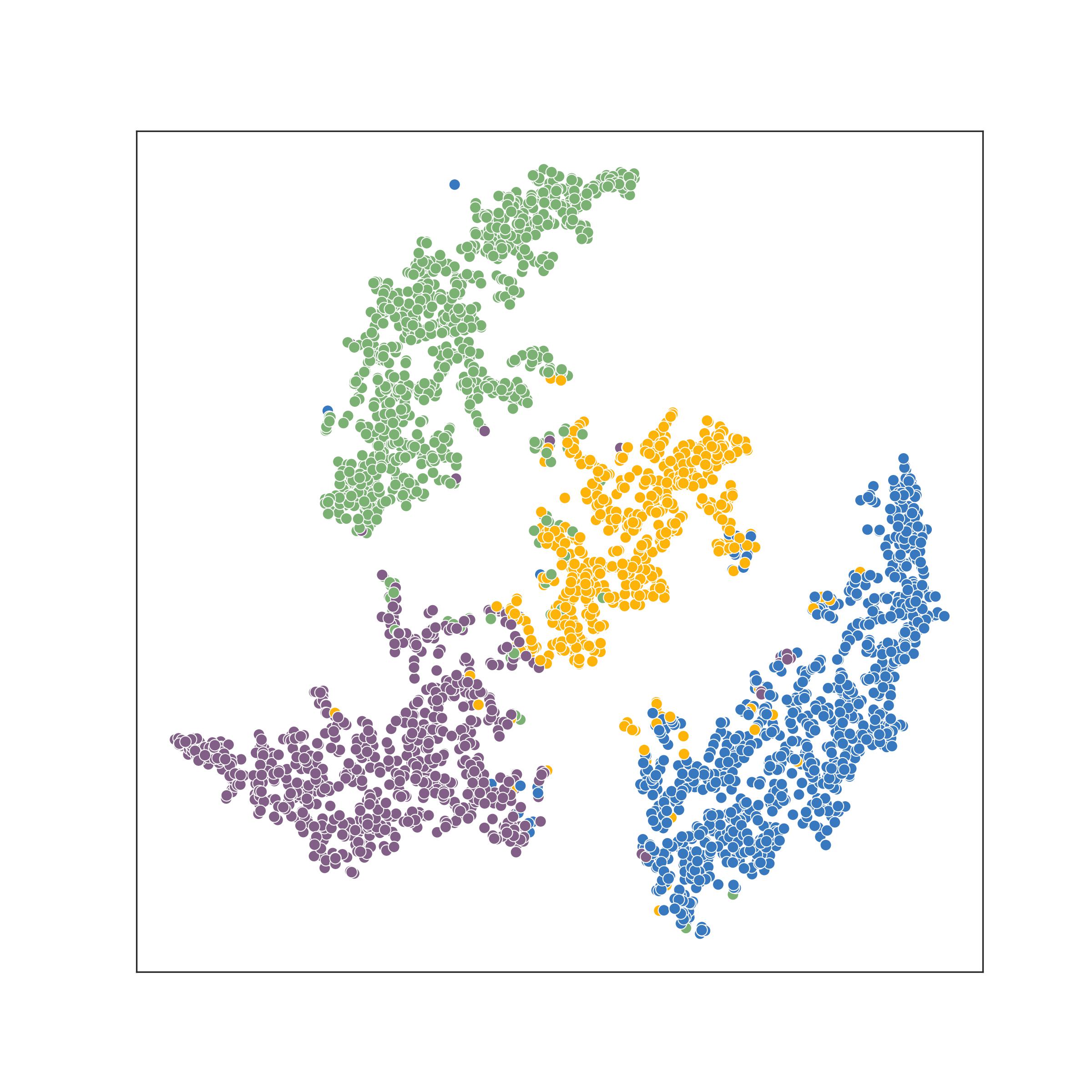}
  }
  \caption{Visualizations of the node representations on DBLP.
  Each data point represents an author, and its color represents its category.}
  \label{fig-visualization}
\end{figure*}

\subsection{Node Representation}

\subsubsection{Node Clustering}
The node clustering task is conducted to validate the effectiveness of the learned node representations.
The representations of testing nodes, which is generated by each trained model, are fed to the K-Means algorithm.
Normalized Mutual Information (NMI) and Adjusted Rand Index (ARI) are employed as the evaluation metrics, where higher scores correspond to better models.
By following \cite{magnn}, we repeat K-Means 10 times for each run of the models, and each model is tested for 10 runs.
As shown in \cref{table-clustering}, our RE-GCN achieves superior performances on all the employed datasets, which demonstrates that our relation embeddings can enable GCN to learn meaningful node representations for heterogeneous graphs.


\subsubsection{Visualizations}

For visual comparisons, we utilize T-SNE \cite{t-sne} to project the learned embeddings of the authors in DBLP into a 2-dimensional space.
As can be observed from \cref{fig-v:gcn}, for the visualization of GCN, authors with the same category tend to be decentralized, and authors with different categories tend to be mixed.
On the contrary, after introducing the relation embeddings to model the heterogeneity, our RE-GCN can effectively learn suitable embeddings.
As shown in \cref{fig-v:regcn}, the authors with identical research interests are more compact, and authors with different research interests are more distinguishable.
Besides, the heterogeneous baseline methods either fail to gather the authors with identical research interests or cannot provide clear boundaries for authors belonging to different categories. 
These results further validate that our RE-GNN can learn effective representations and thus handle the heterogeneous graphs.

\begin{figure}[t]
  \centering
  \subfloat[] { 
    \label{fig-model_size}
    \includegraphics[width=0.90\columnwidth]{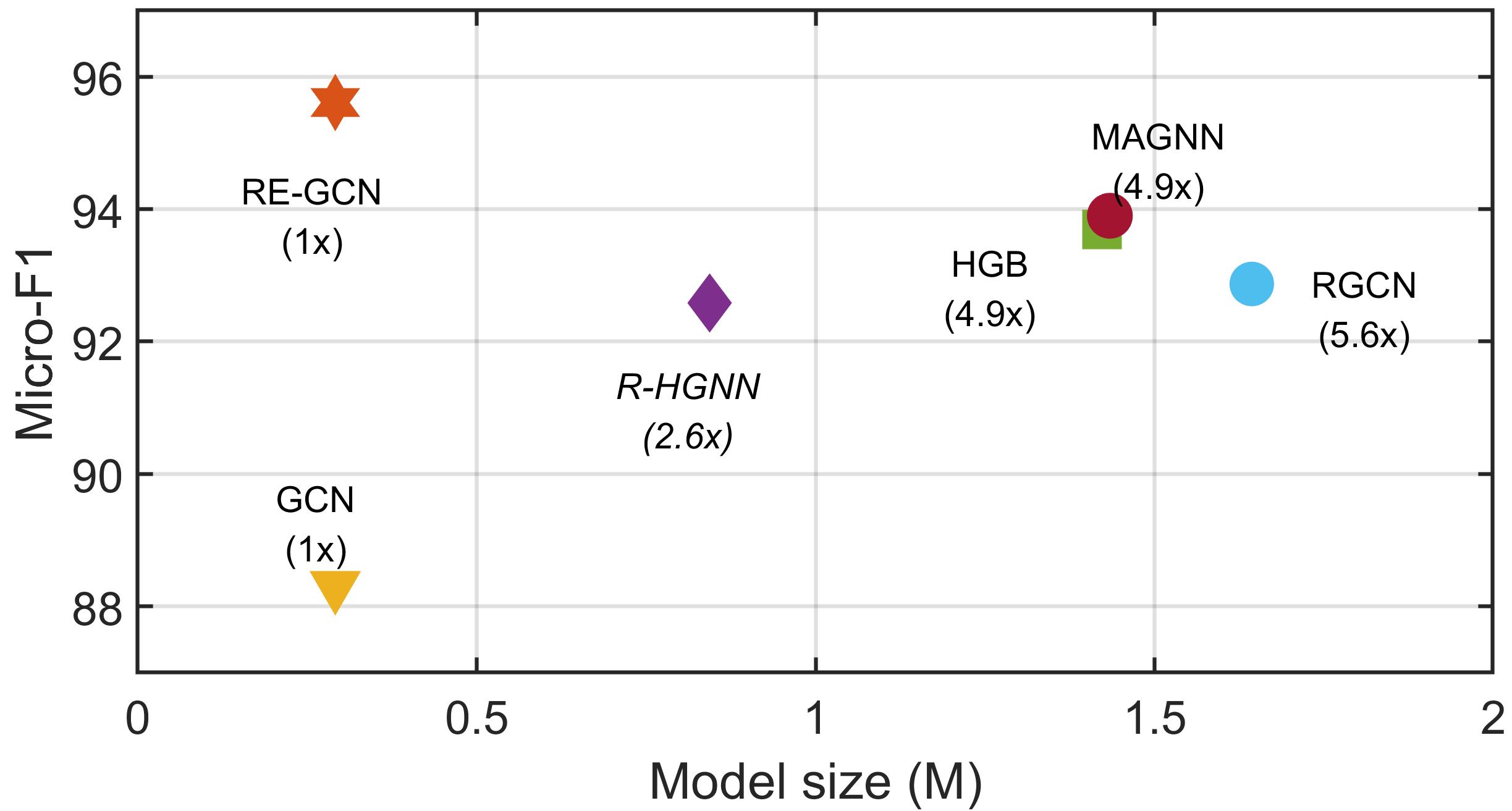}
  }

  \subfloat[] { 
    \label{fig-run_time}
    \includegraphics[width=0.90\columnwidth]{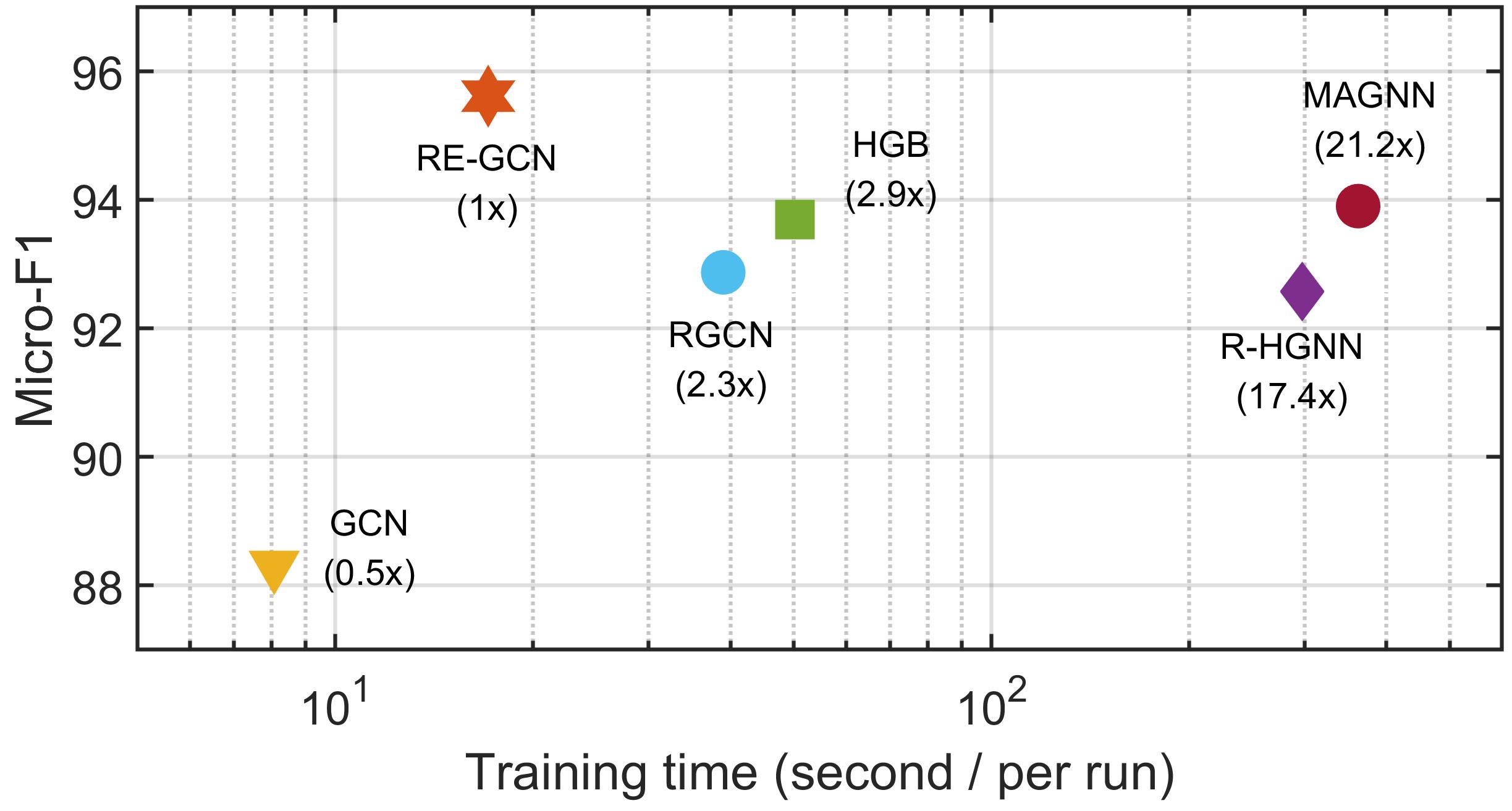}
  }
  \caption{Comparisons of the model size and average training time on DBLP. Note that all the model is trained with 64 hidden units except R-HGNN is trained with 16 hidden units.}
  \label{fig-efficiency}
\end{figure}

\subsection{Efficiency Analysis}
As stated in \cref{sec:method}, our RE-GNN only introduces one parameter for each relation type in each layer compared with the corresponding homogeneous GNN, which is an efficient framework to handle the heterogeneous graphs.
Here, we analyze its experimental efficiency compared to the homogeneous GNNs.
All the experiments are conducted on the Nvidia GeForce RTX 2080Ti GPU with 12 GB memory.
\cref{fig-efficiency} presents the training times and the model sizes for different methods trained on DBLP.
For fair comparisons, all of these methods contain 64 hidden units for each hidden layer.
An exception is R-HGNN, which possesses 16 hidden units (obtained via a hyper-parameter search).

As can be observed, our RE-GCN has an approximately equivalent model size to GCN, which is substantially smaller than the other heterogeneous GNNs.
For example, HGB, a simple heterogeneous GNNs, possesses a 2.6x larger model size than our RE-GCN.
Besides, as can be observed from \cref{fig-run_time}, our RE-GCN outperforms all the heterogeneous GNNs in terms of training speed.
HGB possesses 2.9x training times compared to our RE-GCN.
In addition, two complicated models, i.e., R-HGNN and MAGNN, require more than 17x training times.
In general, compared to these heterogeneous GNNs, our RE-GCN provides the fastest training speed and the fewest model size with the best performances.

\begin{table}[t]
  \centering
  \caption{Performances on OGBN-MAG.}
  \begin{tabular}{c|ccc}
      \toprule
                    & Validation Acc.     & Test Acc.             & Parameters      \\
      \midrule
      MetaPath2vec  & 35.06 ± 0.17        & 35.44 ± 0.36          & 94,479,069\\
      \midrule
      HGT           & 49.89 ± 0.47        & 49.27 ± 0.61          & 21,173,389 \\
      R-GCN         & 49.12 ± 0.50        & 47.89 ± 0.53          & 154,366,772 \\
      R-GSN         & 51.33 ± 0.35        & 50.10 ± 0.42          & 154,373,028 \\
      R-HGNN        & 53.61 ± 0.22        & 52.04 ± 0.26          & 5,638,053 \\ 
      \midrule  
      RE-GCN        & 51.94 ± 0.12        & 50.82 ± 0.18          & \textbf{1,037,171}\\
      RE-GAT        & 53.08 ± 0.26        & \textbf{52.10 ± 0.17} & 1,039,373 \\
      \bottomrule
  \end{tabular}
  \label{table-ogbn-mag}
\end{table}

\subsection{Scalability for Large-scale Graph}

Here, we employ the OGBN-MAG dataset to validate the scalability of our RE-GNN for handling large-scale heterogeneous graphs.
Note that many training techniques can boost the performance on OGBN-MAG, such as multi-stage training \cite{deeperinsights} and adversarial training \cite{flag}, which are not considered in our comparison, because these techniques are orthogonal to model design. 
The results of employed baselines in \cref{table-ogbn-mag}, except for R-GSN, are borrowed from the OGB leaderboards \cite{ogb}.
For R-GSN, we report its official results when removing the FALG training technique \cite{flag}.
For utilizing both GCN and GAT as backbones, we construct 2-layered RE-GNNs followed by a fully connected layer as the output layer.
The hidden dimension is set to 512 for RE-GCN and 64 for RE-GAT with 8 heads.
We utilize the Neighbor Sampling method \cite{graphsage} to train our RE-GNNs with mini-batch.

As can be observed from \cref{table-ogbn-mag}, both RE-GCN and RE-GAT can achieve outstanding performances on OGBN-MAG.
When utilizing GAT as the backbone, our RE-GAT achieves the best performance and outperforms the complicated heterogeneous GNNs, e.g., R-GSN and R-HGNN.
Besides, when utilizing GCN as the backbone, our RE-GCN can still outperform R-GCN.
Meanwhile, our RE-GCN and RE-GAT possess much fewer number of parameters than other heterogeneous methods.
These results demonstrate our RE-GNNs are still effective and efficient on the large-scale heterogeneous graphs.

\begin{figure*}[t]
  \centering
  \subfloat[Layer 1] {
      \label{fig-ew-layer1}
      \includegraphics[height=0.26\columnwidth]{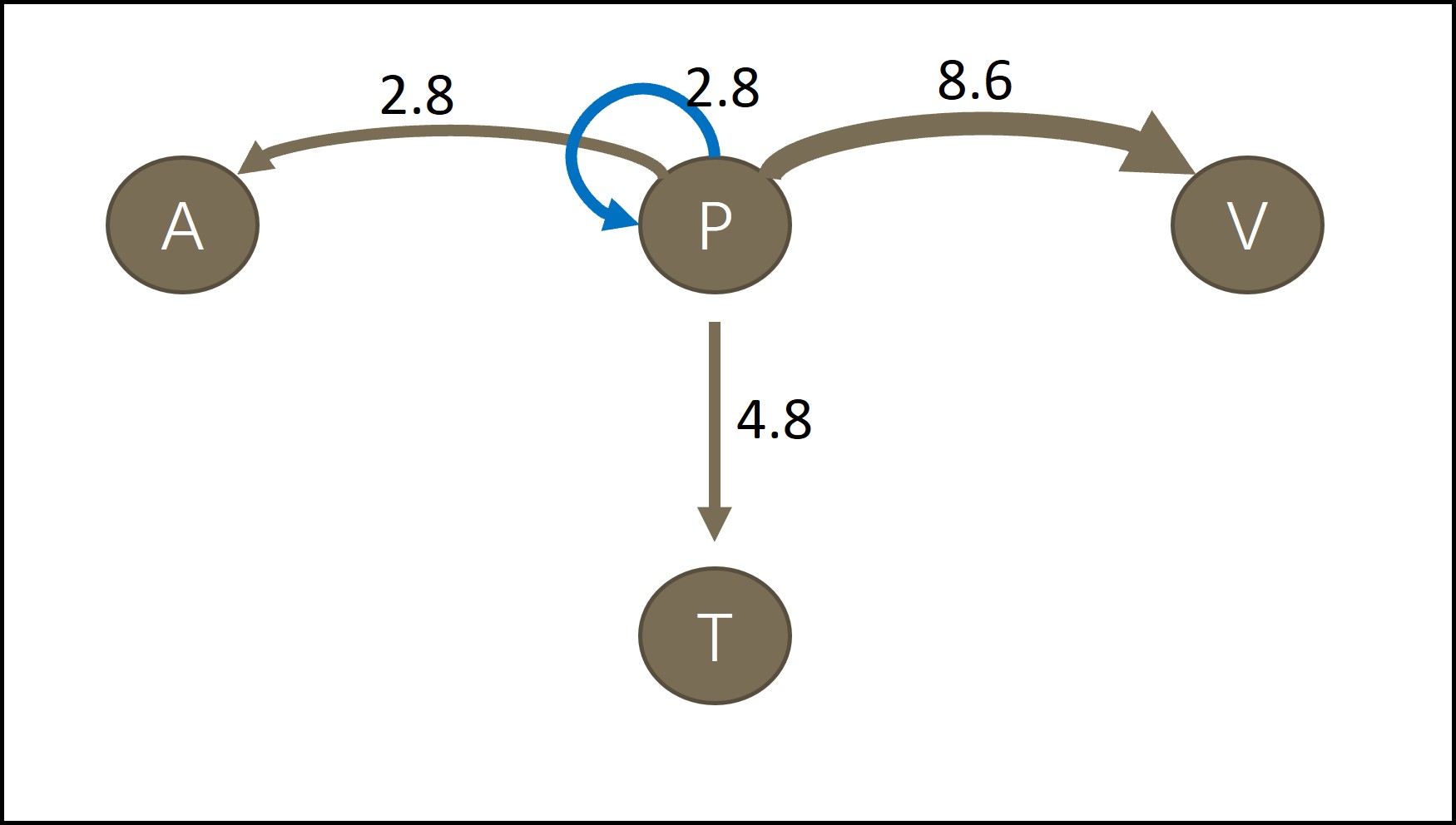}
  }
  \subfloat[Layer 2] { 
      \label{fig-ew-layer2}
      \includegraphics[height=0.26\columnwidth]{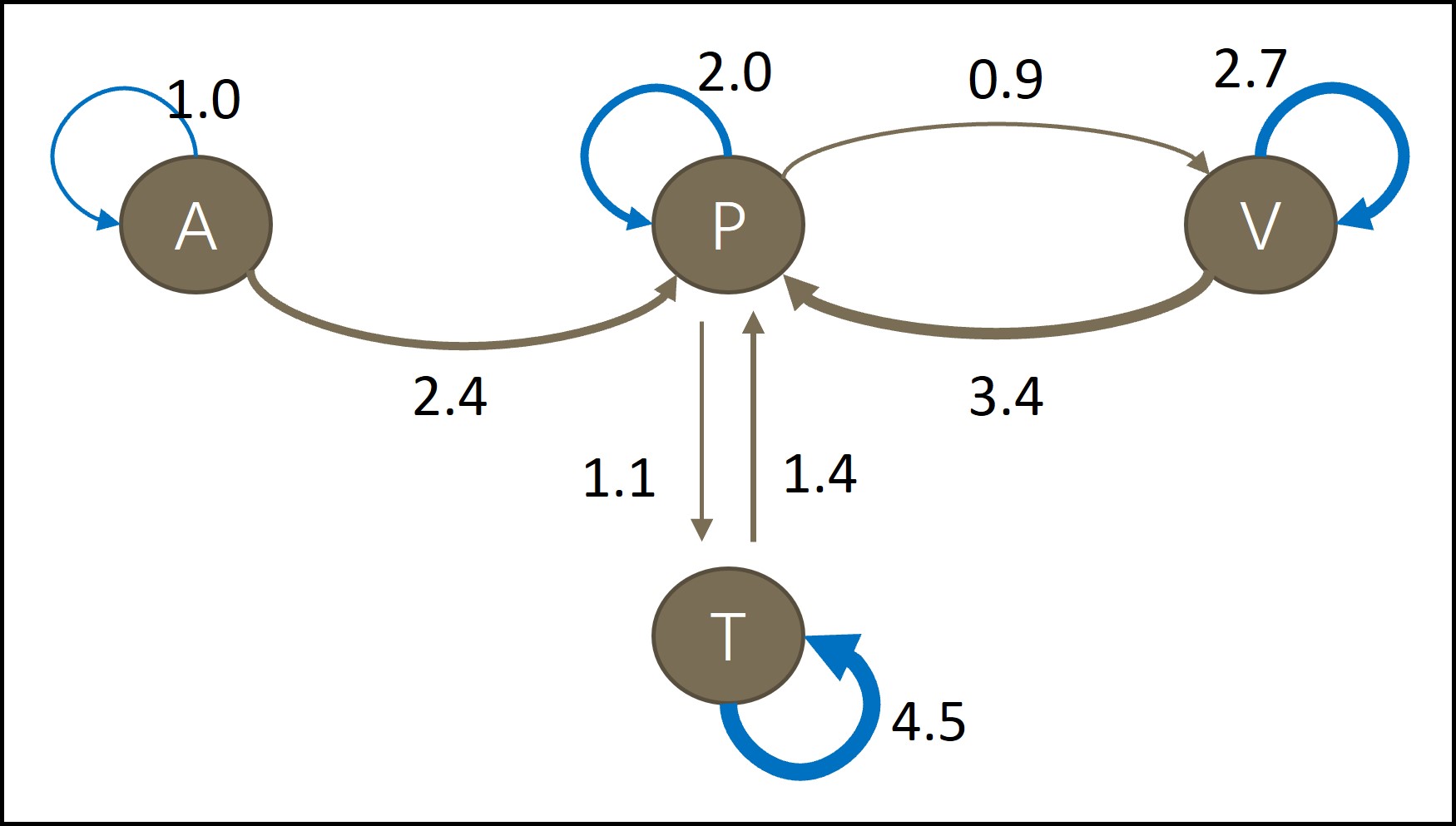}
  }
  \subfloat[Layer 3] { 
      \label{fig-ew-layer3}
      \includegraphics[height=0.26\columnwidth]{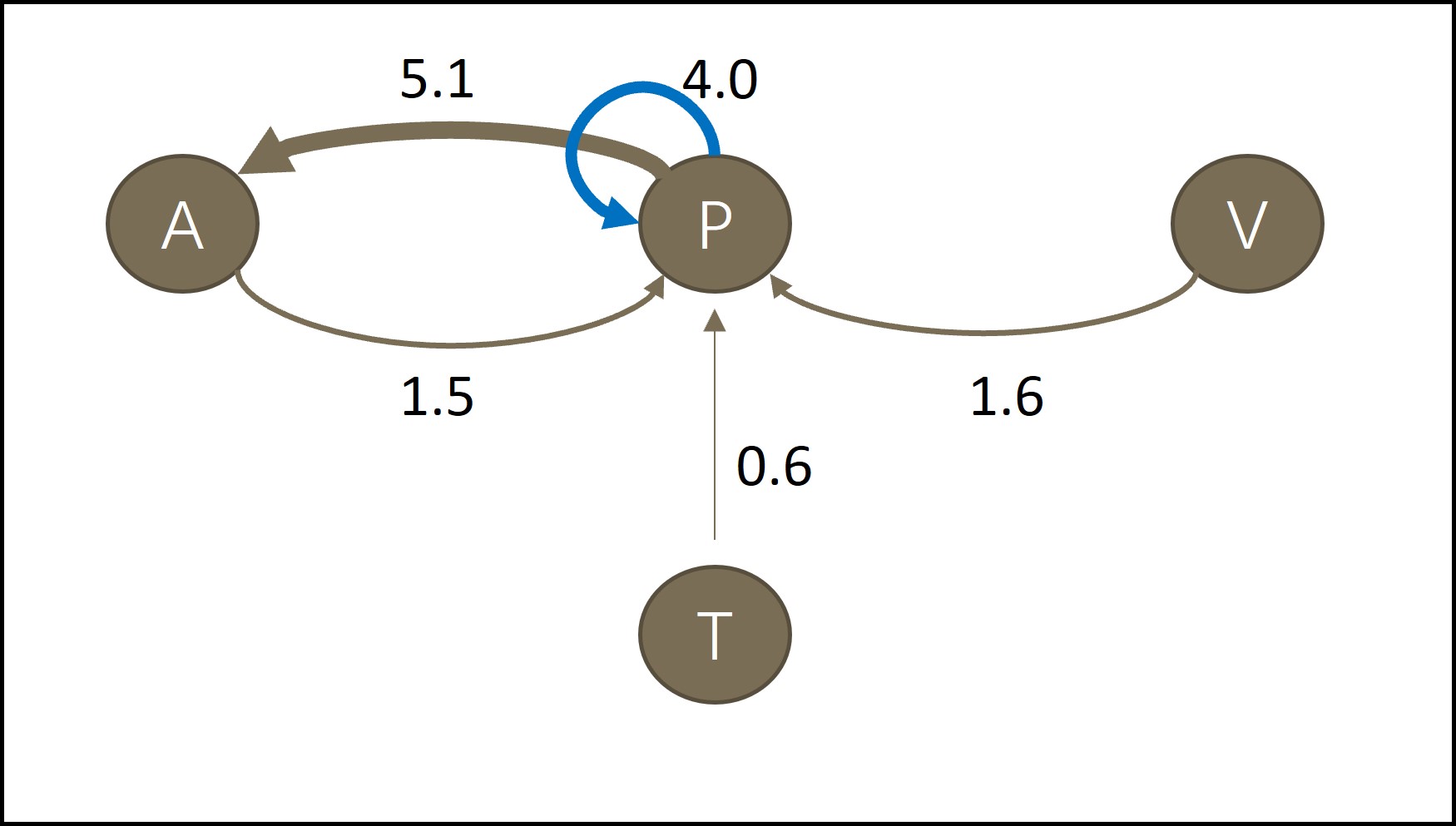}
  }
  \subfloat[Layer 4] { 
      \label{fig-ew-layer4}
      \includegraphics[height=0.26\columnwidth]{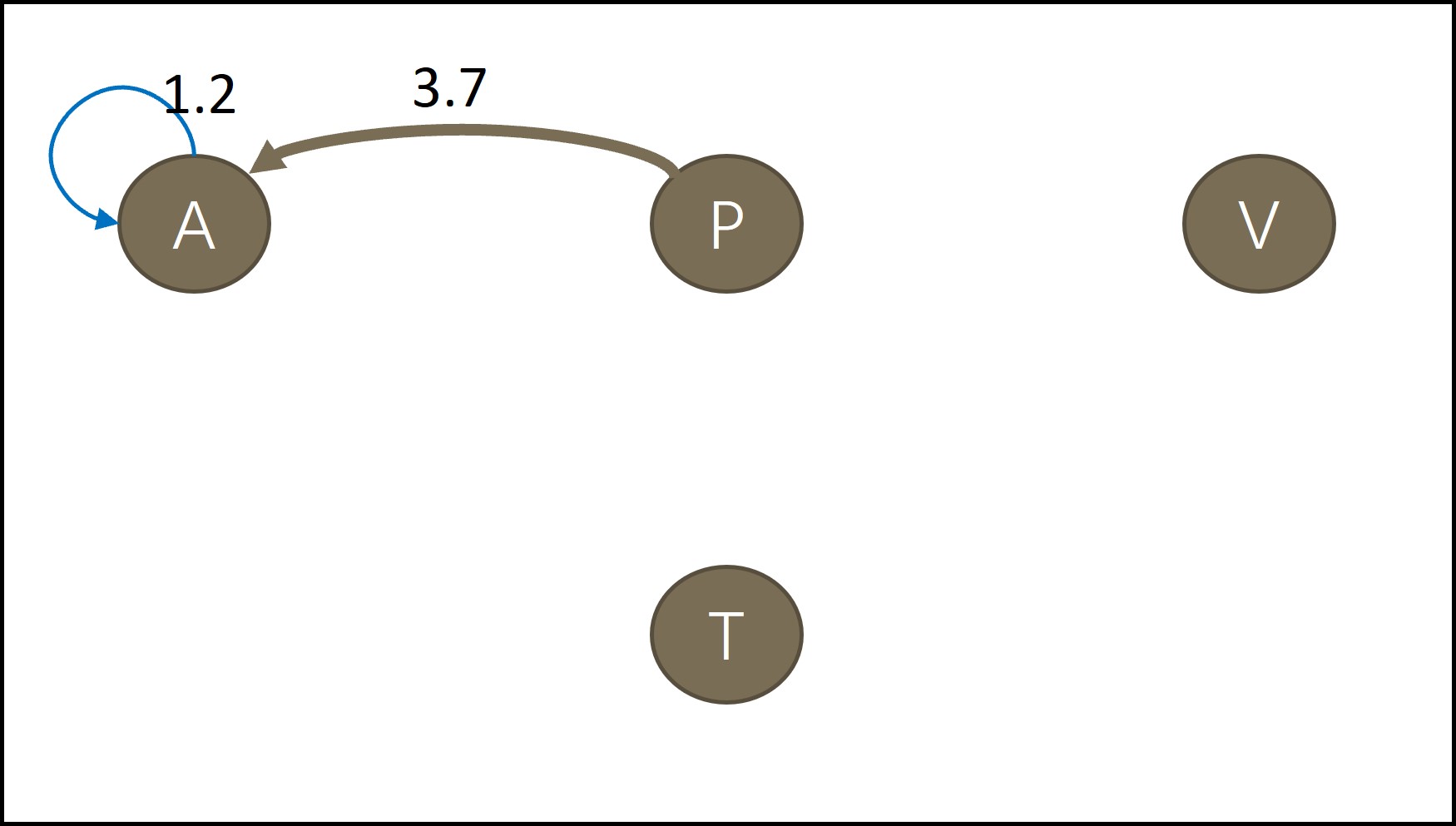}
  }
  \caption{Visualizations of the learned relation embeddings for a 4-layered RE-GCN on the DBLP dataset.
    Note that we remove the relation when its weight is lower than 0.4, for a clear visualization.}
  \label{fig-ew-layer}
\end{figure*}

\subsection{Case Study}
\label{sec:case stduy}
Here, we analyze the learned embeddings of a 4-layered RE-GCN on the DBLP dataset to further study the mechanism of our RE-GNN framework.
As presented in \cref{table-datasets}, DBLP contains four types of nodes and three types of edges.
The task is designed to predict the category of authors.
By considering the self-loop connections and reversing the directed relations, our RE-GNN considers 10 types of relations.

As can be observed from \cref{fig-ew-layer}, the importances of P-* (i.e., P-A, P-T, P-V, and P) relations are much larger than the others in the first layer.
It indicates that RE-GCN believes the paper attributes to be more critical and utilizes them to initialize the features in other types of nodes.
In the second layer, nodes exchange messages according to different types of relations.
Then, in the third layer, there are two groups of dominate messages: 1) P-A; 2) *-P (i.e., A-P, T-P, V-P, and P).
It indicates that the third RE-GCN layer utilizes the messages from papers to update the states of authors, as well as utilizes the messages from other types of nodes to update the states of papers.
In the last layer, the states of authors are updated by the messages from paper nodes and itself.
In general, we can conclude that our RE-GCN can adaptively make the paper nodes to dominate in this case, which is consistent with the common understandings of academic networks.

\begin{table}[t]
  \centering
  \caption{Ablation study on the DBLP dataset.
  Note that \textit{ET emb} represents the edge type relation embeddings, and \textit{SL emb} represents the node-type-specific self-loop embeddings.}
  \begin{tabular}{c|cc|cc}
      \toprule
      & ET emb &SL emb& Macro-F1 & Micro-F1 \\
      \midrule
      GAT & & & 90.19 & 90.94 \\
      GCN & & & 87.39 & 88.31 \\
      \midrule
      GAT-S & & \checkmark & 90.63 & 91.29\\
      GCN-S & &  \checkmark & 88.77 & 89.43\\
      \midrule
      GAT-E & \checkmark & & 94.91 & 95.28 \\
      GCN-E & \checkmark & & 95.15 & 95.48 \\
      \midrule
      RE-GAT & \checkmark & \checkmark & 95.06 & 95.41 \\ 
      RE-GCN & \checkmark & \checkmark & \textbf{95.46} & \textbf{95.80} \\
      \bottomrule
  \end{tabular}
  \label{table-ablation}
  \vspace{-0.4cm}
\end{table}

\subsection{Ablation Study}

Here, we verify the effectiveness of the edge relation embeddings and node-type-specific self-loop embeddings.
Two variants of our RE-GNN are given:
1) GNN-S: homogeneous GNN with node-type-specific self-loop embeddings;
2) GNN-E: homogeneous GNN with edge relation embeddings.

As shown in \cref{table-ablation}, the F1 scores of the GNN-E variants are significantly higher than these of corresponding GNNs.
Meanwhile, the node-type-specific self-loop embeddings give relatively minor improvements.
It indicates that the relation types are more critical than the node-type-specific self-loop types.
Note that the GNN-E variants can also implicitly exploit the general self-loop embeddings, because the normalizations in the aggregation step of GNNs will jointly normalize the message from different types of relations and self-loops.
Besides, RE-GNNs, which explicitly learn the node-type-specific self-loop embeddings, outperform their GNN-E variants.
Therefore, the effectiveness of our node-type-specific self-loop embeddings has also been verified.


\begin{figure*}[t]
  \centering
  \subfloat[$\lambda=0.001$ (Micro-F1: 87.84)] { 
    \label{fig-lambda:0001}
    \includegraphics[width=0.45\columnwidth]{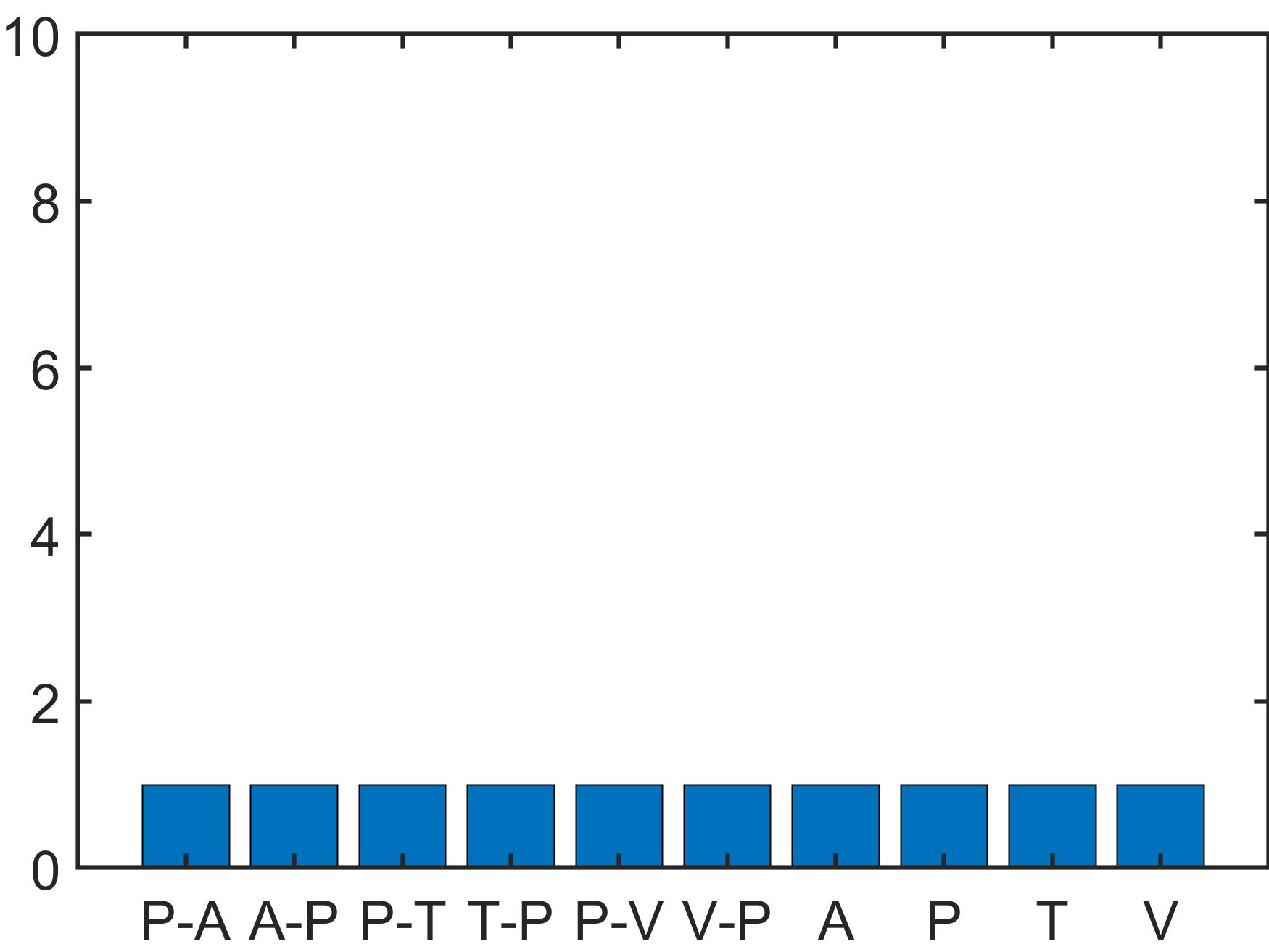}
  }
  \subfloat[$\lambda=1$ (Micro-F1: 90.27)] { 
      \label{fig-lambda:1}
      \includegraphics[width=0.45\columnwidth]{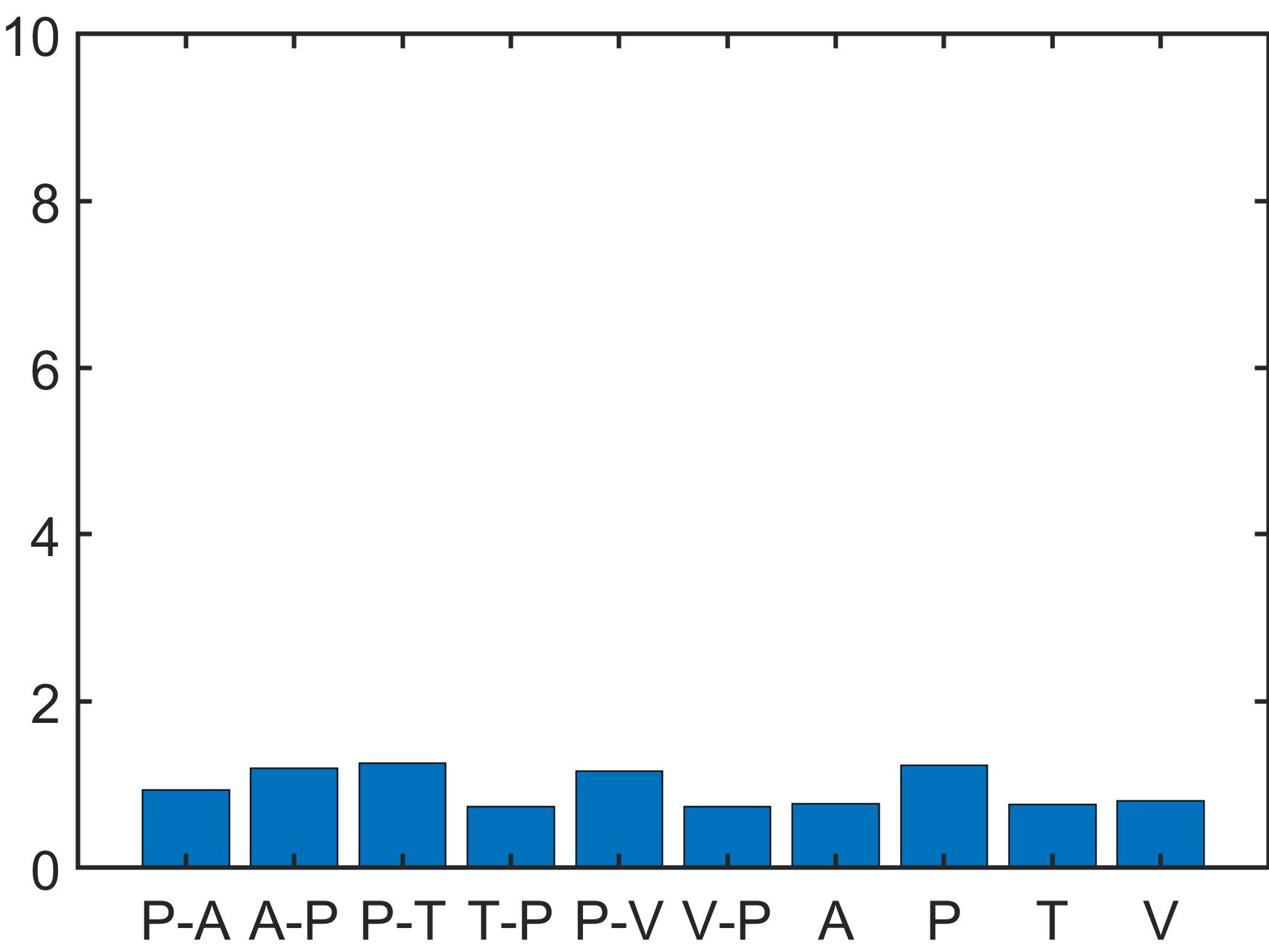}
  }
  \subfloat[$\lambda=100$ (Micro-F1: 95.81)] { 
      \label{fig-lambda:100}
      \includegraphics[width=0.45\columnwidth]{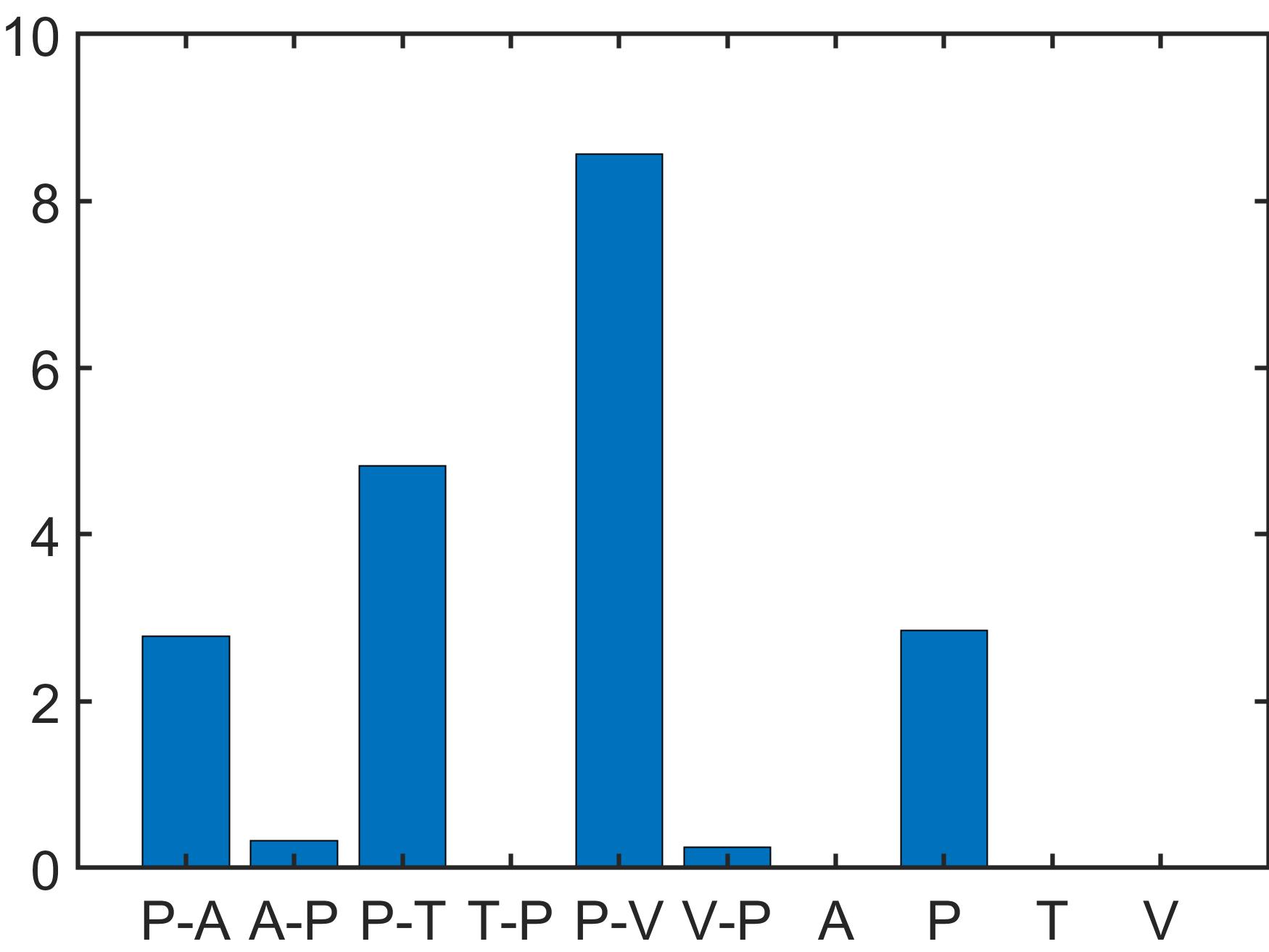}
  }
  \subfloat[$\lambda=1000$ (Micro-F1: 94.84)] { 
      \label{fig-lambda:1000}
      \includegraphics[width=0.45\columnwidth]{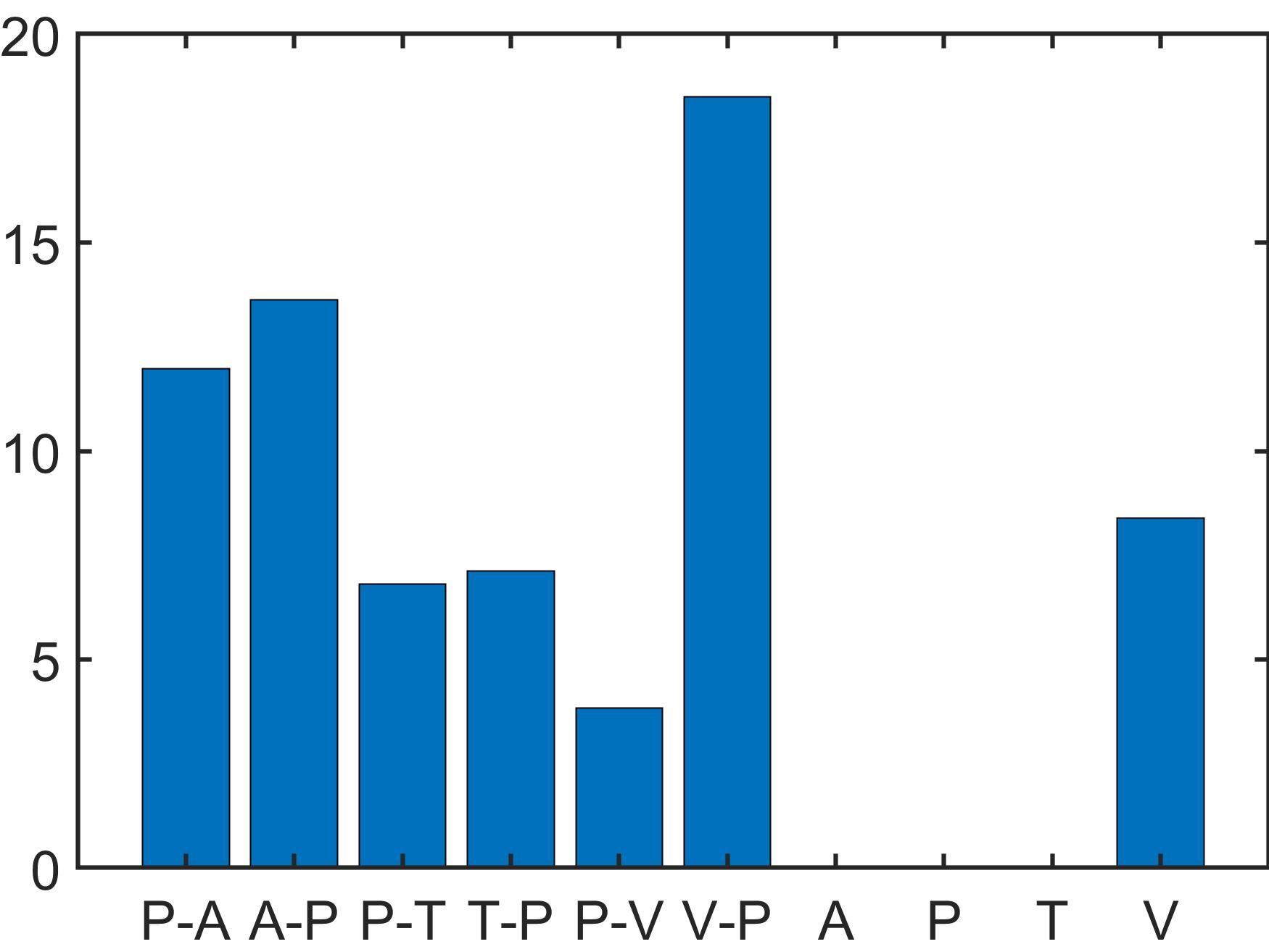}
  }
  \caption{The learned weights of the relations for the first layer of RE-GCN on the DBLP dataset.}
  \label{fig-lambda}
\end{figure*}

\subsection{Impacts of Gradient Scaling Factor}
As stated in \cref{sec-gradient-scaling}, the update of the weights of the relations in each optimization iteration can be scaled via the gradient scaling factor $\lambda$, theoretically.
By using the Adam optimizer, the update of the weights of the relations changes $\lambda$ times.
Here, we experimentally verify the effectiveness of the proposed gradient scaling factor.
\cref{fig-lambda} presents the results of the first layer of RE-GCN, which is learned on the DBLP dataset. 
When $\lambda$ is set to 0.001, the update of the weights of the relations is negligible. 
Under such circumstance, RE-GCN becomes an approximation of GCN with the fixed weights of the relations being ones.
When $\lambda$ is set to 1.0, RE-GCN possesses the original relation embeddings in \cref{original-relation-embeddings,original-selfloop-embeddings}.
As shown in \cref{fig-lambda:1}, the final weights of the relations are similar.
When we set a proper scaling factor, e.g., $\lambda=100$, RE-GCN can learn distinguishable weights of the relations.
However, when $\lambda$ is too large, e.g., 1000, the weights of the relations are very sensitive in the training process and are easy to fall into polarization, i.e., part of the relations are dominating the others.
According to the above results, our gradient scaling factor can help RE-GNN to learn proper relation embeddings and process the heterogeneous graphs.



\section{Conclusion}

This paper proposes a simple yet effective framework, named Relation Embedding based Graph Neural Network (RE-GNN), to assign adequate ability to the homogeneous GNNs for handling the heterogeneous graphs.
Specifically, we exploit only one parameter per relation to model the importance of distinct types of relations and node-type-specific self-loop connections.
To optimize the relation embeddings and the model parameters simultaneously, a gradient scaling factor is proposed to enable the embeddings to converge to appropriate values.
Besides, we interpret the proposed RE-GNN from two perspectives, and theoretically demonstrate that our RE-GCN possesses more expressive power than GTN.
Extensive experiments demonstrate that our RE-GNN can effectively and efficiently handle the heterogeneous graphs and can be applied to various homogeneous GNNs.

\bibliographystyle{IEEEtran}
\bibliography{regnn-trans}

\clearpage

\appendices

\section{Theoretical Impacts of Gradient Scaling Factor}
\label{appendix:scaling-factor}

Here, we discuss the theoretical impacts of the proposed gradients scaling factor, i.e., $\alpha_i = \lambda e_i$, where the relation embedding $e_i$ is scaled by a factor $\lambda > 0$.
The gradient of object function $\mathcal{L}$ to the weight of each relation $\alpha_i$ is $\frac{\partial{\mathcal{L}}}{\partial{\alpha_i}}$.
Then the gradient to each relation embedding $e_i$ is $\lambda \frac{\partial{\mathcal{L}}}{\partial{\alpha_i}}$.
In the original case, where the relation embedding is utilized directly as the weight of the relation ($\alpha_i = e_i$), the gradient to $e_i$ is denoted as $g_i$.
By employing the gradient scaling factor, the scaled gradient to $e_i$ is denoted as $g'_i=\lambda g_i$.

Now, we consider the updating modification for each iteration in the optimization process.
For the gradient descent based optimizers, each relation embedding is updated via $e_i^{next}=e_i-\Delta e_i$, where $\Delta e_i=\kappa(g_i)$ is correlated to the gradient $g_i$.
With the scaling factor $\lambda$, the updating modification becomes $\kappa(\lambda g_i)$.

We argue that $\kappa(\lambda g_i)=\lambda\kappa(g_i)$, for the common gradient optimizers such as SGD, Momentum\cite{momentum}, and Nesterov Momentum \cite{Nesterov}.
For the adaptive optimizers like Adagrad \cite{adagrad} and Adam \cite{adam}, $\kappa(\lambda g_i)=\kappa(g_i)$.
Here, we present the detailed proofs of SGD and Adam optimizers as examples.
The proofs of other commonly utilized optimizers can be generalized by these proofs.

\noindent\textbf{Stochastic Gradient Descent (SGD)}.
For a Stochastic Gradient Descent (SGD) optimizer, a parameter $\theta$ is updated by
\begin{equation}
  \theta_t = \theta_{t-1}-\eta g_t,
  \tag{S1}
\end{equation}
where $\eta$ stands for the learning rate and $g_t=\frac{\partial \mathcal{L}}{\partial \theta_{t-1}}$ represents the gradient of a batch of the input data.
Thus, the updating modification function is $\kappa(g) = -\eta g$.
Then, $\kappa(g')=\kappa(\lambda g) = -\eta \lambda g = \lambda (-\eta g) = \lambda\kappa(g)$.

\noindent\textbf{Adaptive Moment Estimation (Adam)}.
Adaptive Moment Estimation (Adam) is a typical gradient descent method which computes an adaptive learning rate for each parameter.
In each iteration $t$, it firstly computes the exponential averages of the gradient $m_t$ and the squared gradient $v_t$ respectively as
\begin{equation}
  m_t = \beta_1 m_{t-1} + (1-\beta_1) g_t,
  \tag{S2}
\end{equation}
\begin{equation}
  v_{t} = \beta_2 v_{t-1} + (1-\beta_2) g_t^2.
  \tag{S3}
\end{equation}
where $\beta_1$ and $\beta_2$ are the two pre-defined hyperparameters.
Both $m_{t}$ and $v_{t}$ are initialized as vectors of zeros.
Then, the bias elimination is utilized to obtain
\begin{equation}
  \hat{m}_{t} = \frac{m_{t}}{1-\beta_1^t},
  \tag{S4}
\end{equation}
\begin{equation}
  \hat{v}_{t} = \frac{v_{t}}{1-\beta_2^t}.
  \tag{S5}
\end{equation}
At last, the parameter is updated accordingly.
\begin{equation}
  \theta_{t} = \theta_{t-1}-\eta \frac{\hat{m}_{t}}{\sqrt{\hat{v}_{t}}+\epsilon},
  \tag{S6}
\end{equation}
where $\epsilon>0$ is an extremely small number which can be neglected when $v_{t}\not=0$.
Then, the updating modification function is $\kappa(g_t) = -\eta \frac{\hat{m}_{t}}{\sqrt{\hat{v}_{t}}+\epsilon}$.
In the following proof, induction is employed to prove $\kappa(\lambda g_t) = \kappa(g_t)$.
\begin{proof}
  Initially, we have $m_0 = 0$ and $v_0 = 0$.
  Let $g'_t=\lambda g_t$.
  Then, $\kappa(\lambda g) = \kappa(g')$.
  
  In the first iteration, since $m_0=v_0=0$, it is easy to validate $m'_1 = \lambda m_1$, $v'_{1} = \lambda^2 v_{1}$, $\hat{m}'_{1} = \lambda \hat{m}_1$ and  $\hat{v}'_1 = \lambda^2 \hat{v}_1$.
  Then, the updating function is $\kappa(\lambda g_{t}) =-\eta \frac{\lambda \hat{m}_{1}}{\sqrt{\lambda^2\hat{v}_{1}}+\epsilon} =\kappa(g_{t})$.
  Since $\epsilon$ is a very small number, which is only utilized to prevent a zero denominator in practice, we ignore its effect here.

  Assuming that in the $(t-1)$-th iteration, we have obtained $\hat{m}'_{t-1} = \lambda \hat{m}_{t-1}$, $\hat{v}'_{t-1} = \lambda^2 \hat{v}_{t-1}$ and $\kappa(g'_{t-1}) = \kappa(g_{t-1})$.
  Then, in the $t$-th iteration, we can obtain
  \begin{equation}
    \begin{split}
      m'_t &= \beta_1 m'_{t-1} + (1-\beta_1) g'_t \\
         &= \beta_1 \lambda m_{t-1} + (1-\beta_1) \lambda g_t \\
         &=\lambda (\beta_1 m_{t-1} + (1-\beta_1) g_t) \\
         &=\lambda m_t.
      \end{split}
      \tag{S7}
  \end{equation}
  Similarly, we can obtain 
  \begin{equation}
    \begin{split}
      v'_t &= \beta_2 v'_{t-1} + (1-\beta_2) g'^2_t \\
           &= \beta_2 \lambda^2 v_{t-1} + (1-\beta_2) \lambda^2 g_t^2 \\
           &=\lambda^2 (\beta_2 m_{t-1} + (1-\beta_2) g_t^2) \\
           &=\lambda^2 v_t.
    \end{split}
    \tag{S8}
  \end{equation}
  Then, for the bias eliminated $\hat{m}'_t$ and $\hat{v}'_t$, we can compute $\hat{m}'_t=\lambda\hat{m}_t$ and $\hat{v}'_t=\lambda^2\hat{v}_t$ respectively.
  At last, the updating modification is $\kappa(\lambda g_{t}) =-\eta \frac{\hat{m}'_{1}}{\sqrt{\hat{v}'_{1}}+\epsilon} =\kappa(g_{t})$.
\end{proof}

Then, we consider the modification for the weight of each relation $\Delta \alpha_i$.
Since, $\alpha_i$ is not a parameter, its modification is correlated to $\Delta e_i$.
In the original case, $\Delta \alpha= \Delta e_i = \kappa(g_i)$.
By employing the gradient scaling factor, the modification of the weight of each relation is $\Delta \alpha'=\lambda \Delta e_i = \lambda \kappa(g'_i)$.
As stated above, for the common gradient optimizers, such as SGD, Momentum and Nesterov, $\Delta \alpha'=\lambda^2 \Delta \alpha$.
For the adaptive optimizers like Adagrad and Adam, $\Delta \alpha'=\lambda \Delta \alpha$.

\section{Proof of Lemma 3}

\begin{proof}

$\forall h \in \mathcal{H}_{in}$, we can obtain
\begin{equation}
  (f_1\circ f_2)(h) = \sigma(\sigma(hW_1+b_1)W_2+b_2),
  \tag{S10}
  \label{eq-2-layered-fc}
\end{equation}
where $\sigma$ is a ReLU function, $h$, $W_1$, and $W_2$ are bounded by $k$, $k_{w_1}$, $k_{w_2}$, respectively.
By letting $t=hW_1$, we can obtain $t_i=hw_i^T$.
Since $hh^T=\sum_{j}h_j^2<k$ and $w_iw_i^T=\sum_{j}w_{ij}^2<k_{w_1}$, we can calculate $|t_i|=|hw_i^T|=|\sum_{j}h_jw_{ij}|\leq\sum_{j}|h_jw_{ij}|\leq \frac{1}{2}\sum_{j}(h_j^2+w_{ij}^2)=\frac{k+k_{w_1}}{2}$.
Then, there is a $|b_i|\leq \frac{k+k_{w_1}}{2}$, when $b_i+t_i\geq 0$, i.e., there exists a $b_{1}'$ with a bound of $\frac{k+k_1}{2}$, when $hW_1+b_{1}'>0$.
Thus, Eq. \eqref{eq-2-layered-fc} can be rewritten as
\begin{equation}
  (f_1\circ f_2)(h) = \sigma(hW_1W_2+b_{1}'W_2+b_2).
  \tag{S11}
\end{equation}
Then, when $W_1W_2 = W$ and $b_{1}'W_2+b_2=b$, $(f_1\circ f_2)$ equals to $f$.
There exists a simple solution to the above formulas, i.e., $W_1=W, W_2=I, b_2=b-b_{1}'$.
  
\end{proof}

\section{Proof of Lemma 4}

\begin{proof}
  According to the definition, the bound of $o$ is 
  \begin{equation}
    ||o||_2^2 = \sum_{j}o_j^2 =\sum_{j}\left(\sum_{i}p_{i}h_{ij}\right)^2.
    \label{eq-S4-1}
    \tag{S12}
  \end{equation}
  Since $\sum_{i=1}^r p_i=1$, and $(\cdot)^2$ is a convex function,
  \begin{equation}
    \sum_{j}\left(\sum_{i}p_{i}h_{ij}\right)^2 \leq \sum_{j}\left(\sum_{i}p_{i}h_{ij}^2\right),
    \tag{S13}
  \end{equation}
  according to the Jensen’s inequality.
  Then,
  \begin{equation}
    \begin{split}
      \sum_{j}\left(\sum_{i}p_{i}h_{ij}^2\right) &= \sum_{i}p_{i}\left(\sum_{j}h_{ij}^2\right)\\
                    & < \sum_{i}p_{i} k\\
                    & = k.
    \end{split}
    \tag{S14}
  \end{equation}
  At last, we can obtain $||o||_2^2<k$.
\end{proof}

\section{Proof of Corollary 5}
\begin{proof}

  For Eq. \eqref{eq-2-layered-REGCN-2}, X is the collection of $x$.
  Each row of $X$, which is denoted as $x_i$, is bounded by $\xi$.
  $\tilde{A}=\hat{D}^{-1}\hat{A}$ is a non-negative matrix, where the summation of each row equals to $1$, i.e., $\sum_{j}\tilde{a}_{ij}=1$ and $\tilde{a}_{ij} \geq 0$.
  Let $H=\tilde{A}X$. 
  According to Lemma 4, the bound of each row $h_i$ is $\xi$.

  Then, according to Lemma 3, we can conclude that there exists a $b'$ with a bound of $\frac{k+k_1}{2}$, $hW_1+b_{1}'>0$, i.e., $HW_1+B_1'>0$.
  Thus, Eq. \eqref{eq-2-layered-REGCN-2} can be rewritten as
  \begin{multline}
    Z_H = \sigma(\tilde{A}^{(1)}_H\tilde{A}^{(0)}_HXW^{(0)}_HW^{(1)}_H
    +\tilde{A}^{(1)}_HB^{(0)'}_HW^{(1)}_H+B^{(1)}_H).
    \tag{S17}
    \label{eq-2-lengthed-resgc}
  \end{multline}
  Since $\tilde{A}^{(1)}_H$ is a matrix, which is normalized in each row, and $B^{(0)'}_H$ is a column equivalent matrix, we can obtain 
  \begin{equation}
    \tilde{A}^{(1)}_HB^{(0)'}_H=B^{(0)'}_H.
    \tag{S18}
    \label{S18}
  \end{equation}
  Besides,
  \begin{equation}
    \begin{split}
      \tilde{A}^{(1)}_H\tilde{A}^{(0)}_H &= \hat{D}^{(1)}\hat{A}^{(1)}_H\hat{D}^{(0)}\hat{A}^{(0)}_H\\
      &=\hat{D}^{(1)}\hat{D}^{(0)}\hat{A}^{(1)}_H\hat{A}^{(0)}_H\\
      & = \hat{D}_{H_2}\hat{A}_{H2}\\ 
      &= \tilde{A}_{H2}.
    \end{split}
    \tag{S19}
    \label{S19}
  \end{equation}
  With Eqs. \eqref{S18} and \eqref{S19}, Eq. \eqref{eq-2-lengthed-resgc} can be reformed to
  \begin{equation}
    Z_H = \sigma\left(\tilde{A}_{H2}XW^{(0)}_HW^{(1)}_H+B^{(0)'}_HW^{(1)}_H+B^{(1)}_H\right).
    \tag{S20}
  \end{equation}
  Similar to the proof of Lemma 3, we can easily construct a solution that $W_H^{(0)} = W_P$, $W_H^{(1)} =I$, $B_H^{(1)}=B_P-B_H^{(0)'}$.

\end{proof}

\section{Proof of Theorem 6}

\begin{proof}
  Theorem 1 can be proved in two steps.
  \begin{itemize}
    \item T1(1). For any $L$-lengthed one-layered GTN, there exists an $L$-layered RE-GCN which is equivalent to it.
    \item T1(2). For any $L$-lengthed $K$-layered GTN, there exists an $(LK)$-layered RE-GCN which is equivalent to it, where $K>1$.
  \end{itemize}

  According to Corollary 5, for an $L$-lengthed GTN layer, we can also obtain a stack of $L$ RE-GCN layers which is equivalent to it.
  This equivalency can be achieved via removing the ReLU function and choosing a proper bias vector $b^{(L)}$ in each layer (except the last layer).
  Note that the bias vector $b^{(l)}$ is bounded by $\frac{k_{in}^{(l)}+k_{w}^{(l)}}{2}$.
  $k^{(l)}_{in}$ and $k_{w}^{(l)}$ are the bounds of the input features and parameters in the $L$-th layer, respectively.
  Therefore, T1(1) is proved.
  
  For T1(2), we can employ a composite of $L$ RE-GCN layers which is equivalent to each $L$-lengthed GTN layer.
  Then, we can stack these $(KL)$ RE-GCN layers, which is equivalent to the $L$-lengthed $K$-layered GTN.
  Therefore, T1(2) is proved.
\end{proof}

\section{Proof of Theorem 7}

\begin{proof}

  Theorem 2 can be proved in two steps.
  \begin{itemize}
    \item T2(1). There exists a 2-layered RE-GCN, which cannot be represented by any $L$-lengthed one-layered GTN.
    \item T2(2). There exists a 2-layered RE-GCN, which cannot be represented by any $L$-lengthed $K$-layered GTN, where $K>1$.
  \end{itemize}
  
  Consider a graph with only one node and a self-loop connection.
  Then, a GCN layer is degenerated to an MLP layer.
  Since the composite of two MLP layers can be a non-linear mapping while one MLP layer can only be a linear mapping (by ignoring the last ReLU function), there exists a composite of two MLP layers that one MLP layer cannot be equivalent to.
  Therefore, T2(1) is proved.

  The adjacency matrix enables the nodes (samples) to exchange messages with others.
  Then, the effects of neighbourhood aggregation with $A$ cannot be replaced by the effects of weight projection with $W$.
  For a $K$-layered GTN ($K>1$) with an arbitrary learned adjacency matrix $A_P$, it should satisfy that $A_P^{k_1}$ = $A_{H_1}$, $A_P^{k_2}$ = $A_{H_2}$, where $k_1+k_2=K$.
  However, this cannot be satisfied at all the time for any possible $A_{H_1}$ and $A_{H_2}$.
  For example, if $A_{H_1}$ is a non-singular matrix, while $A_{H_2}$ is a singular matrix, no proper $A_P$ can be obtained.
  The reason is that $A^n$ is a non-singular matrix, if and only if the matrix $A$ is non-singular.
  Therefore, T2(1) is proved.

  
\end{proof}

\end{document}